\documentclass[preprint,review,12pt]{elsarticle}

\usepackage{amssymb}

\usepackage{setspace}
\usepackage{caption}
\usepackage{subcaption}
\usepackage{amsmath}
\usepackage{systeme}
\usepackage{url}
\usepackage{xcolor}
\usepackage[linesnumbered,ruled,vlined]{algorithm2e}

\SetCommentSty{mycommfont}
\SetKwInput{KwInput}{Input}                
\SetKwInput{KwOutput}{Output}              
\SetKwProg{Fn}{Function}{}{}

\journal{Array}

\begin{document}

\captionsetup[figure]{labelfont={bf},labelformat={default},labelsep=period,name={Fig.}}
\captionsetup[table]{labelfont={bf},labelformat={default},labelsep=period,name={Table}}
\begin{frontmatter}

\title{Random projection tree similarity metric for SpectralNet}

\author[1]{Mashaan Alshammari\corref{cor1}}
\ead{mashaan.awad1930@alum.kfupm.edu.sa}
\author[2]{John Stavrakakis}
\ead{john.stavrakakis@sydney.edu.au}
\author[3]{Adel F. Ahmed}
\ead{adelahmed@kfupm.edu.sa}
\author[2]{Masahiro Takatsuka}
\ead{masa.takatsuka@sydney.edu.au}

\cortext[cor1]{Corresponding author}

\address[1]{Independent Researcher, Riyadh, Saudi Arabia}
\address[2]{School of Computer Science, The University of Sydney, NSW, Australia}
\address[3]{Information and Computer Science Department, King Fahd University of Petroleum and Minerals, Dhahran, Saudi Arabia}

\begin{abstract}
\begin{singlespace}
SpectralNet is a graph clustering method that uses neural network to find an embedding that separates the data. So far it was only used with $k$-nn graphs, which are usually constructed using a distance metric (e.g., Euclidean distance). $k$-nn graphs restrict the points to have a fixed number of neighbors regardless of the local statistics around them. We proposed a new SpectralNet similarity metric based on random projection trees (rpTrees). Our experiments revealed that SpectralNet produces better clustering accuracy using rpTree similarity metric compared to $k$-nn graph with a distance metric. Also, we found out that rpTree parameters do not affect the clustering accuracy. These parameters include the leaf size and the selection of projection direction. It is computationally efficient to keep the leaf size in order of $\log(n)$, and project the points onto a random direction instead of trying to find the direction with the maximum dispersion.
\end{singlespace}
\end{abstract}


\end{frontmatter}




\section{Introduction}\label{Introduction}
Graph clustering is one of the fundamental tasks in unsupervised learning. The flexibility of modeling any problem as a graph has made graph clustering very popular. Extracting clusters' information from graph is computationally expensive, as it usually done via eigen decomposition in a method known as spectral clustering. A recently proposed method, named as SpectralNet \cite{shaham2018spectralnet}, was able to detect clusters in a graph without passing through the expensive step of eigen decomposition.

SpectralNet starts by learning pairwise similarities between data points using Siamese nets \cite{Bromley1993Signature}. The pairwise similarities are stored in an affinity matrix $A$, which is then passed through a deep network to learn an embedding space. In that embedding space, pairs with large similarities fall in a close proximity to each other. Then, similar points can be clustered together by running $k$-means in that embedding space. In order for SpectralNet to produce accurate results, it needs an affinity matrix with rich information about the clusters. Ideally, a pair of points in the same cluster should be connected with an edge carrying a large weight. If the pair belong to different clusters, they should be connected with an edge carrying a small weight, or no weight which is indicated by a zero entry in the affinity matrix. 

SpectralNet uses Siamese nets to learn informative weights that ensure good clustering results. However, the Siamese nets need some information beforehand. They need some pairs to be labelled as negative and positive pairs. Negative label indicates a pair of points belonging to different clusters, and a positive label indicates a pair of points in the same cluster. Obtaining negative and positive pairs can be done in a semi-supervised or unsupervised manner. The authors of SpectralNet have implemented it as a semi-supervised and an unsupervised method. Using the ground-truth labels to assign negative and positive labels, makes the SpectralNet semi-supervised. On the other hand, using a distance metric to label closer points as positive pairs and farther points as negative pairs, makes the SpectralNet unsupervised. In this study, we are only interested in an unsupervised SpectralNet.

\begin{figure}[h]
	\centering
	\includegraphics[width=0.75\textwidth,height=20cm,keepaspectratio]{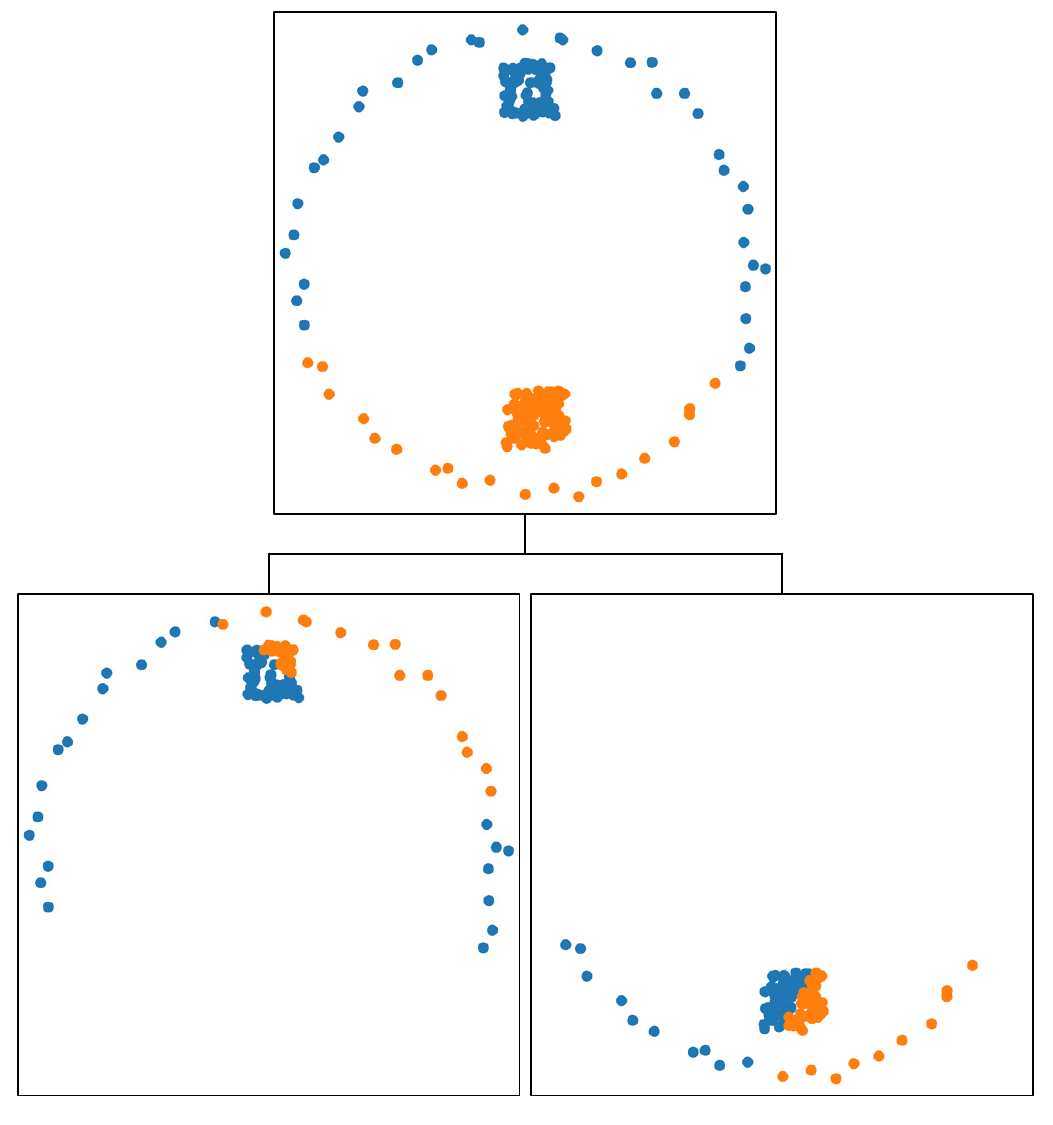}	
	\caption{An example of rpTree; points in blue are placed in the left branch and points in orange are placed in the right branch. (Best viewed in color)}
	\label{Fig:rpTree}
\end{figure}

Unsupervised SpectralNet uses a distance metric to assign positive and negative pairs. A common approach is to get the nearest $k$ neighbors for each point and assign those neighbors as positive pairs. A random selection of farther points are labeled as negative pairs. But this approach restricts all points to have a fixed number of positive pairs, which is unsuitable if clusters have different densities. In this work, we proposed a similarity metric based on random projection trees (rpTrees) \cite{Dasgupta2008Random,Freund2008Learning}. An example of an rpTree is shown in Fig.\ \ref{Fig:rpTree}. rpTrees do not restrict the number of positive pairs, as this depends on how many points in the leaf node.

The main contributions of this work can be summarized in the following points:
\begin{itemize}
	\item Proposing a similarity metric for SpectralNet based on random projection trees (rpTrees) that does not restrict the number of positive pairs and produces better clustering accuracy.
	
	\item Investigating the influence of the leaf size parameter $n_0$ on the clustering accuracy.
	
	\item Performing an in-depth analysis of the projection direction in rpTrees, and examine how it influences the clustering accuracy of SpectralNet.
\end{itemize}



\section{Related work}
\label{RelatedWork}

\subsection{Graph neural networks (GNNs)}
GNNs became researchers’ go-to option to perform graph representation learning. Due to its capability in fusing nodes’ attributes and graph structure, GNN has been widely used in many applications such as knowledge tracing \cite{Song2022Survey} and sentiment analysis \cite{ZHOU2020Modeling}. The most well-known form of GNN is graph convolutional network (GCN) \cite{kipf2017semi}. 

Researchers have been working on improving GCN. Franceschi et al. \cite{franceschi2019learning} have proposed to learn the adjacency matrix $A$ by running GCN for multiple iterations and adjusting the graph edges in $A$ accordingly. Another problem with GCN is its vulnerability to adversarial attack. Yang et al. used GCN with domain adaptive learning \cite{Yang2022Robust}. Domain adaptive learning attempts to transfer the knowledge from a labeled source graph to unlabeled target graph. Unseen nodes from the target graph can later be used for node classification.

\subsection{Graph clustering using deep networks}
GCN performs semi-supervised node classification. Due to limited availability of labeled data in some applications, researchers developed graph clustering using deep networks. Yang et al. developed a deep model for network clustering \cite{Yang2021Variational}. They used graph neural network (GCN) to encode the adjacency matrix $A$ and the feature matrix $X$. They also used multilayer perceptron (MLP) to encode the feature matrix $X$. The output is clustered using Gaussian mixture model (GMM), where GMM parameters are updated throughout training. A similar approach was used by Wang et al. \cite{WANG2022Learning}, where they used autoencoders to learn latent representation. Then, they deploy the manifold learning technique UMAP \cite{McInnes2018UMAP} to find a low dimensional space. The final clustering assignments are given by $k$-means. Affeldt et al. used autoencoders to obtain $m$ representations of the input data \cite{Affeldt2020Spectral}. The affinity matrices of these $m$ representations are merged into a single matrix. Then spectral clustering was performed on the merged matrix. One drawback with this approach is that it still needs eigen decomposition to find the embedding space.

SpectralNet is another approach for graph clustering using deep networks, which was proposed by Shaham et al. \cite{shaham2018spectralnet}. They used Siamese nets to construct the adjacency matrix $A$, which is then passed through a deep network. Nodes in the embedding space can be clustered using $k$-means. An extension to SpectralNet was proposed by \citet{Huang2019MultiSpectralNet}, where multiple Siamese nets are trained on multiple views. Each view is passed into a neural network to find an embedding space. All embedding spaces are fused in the final stage, and $k$-means was run to find the cluster labels. Another approach to employ deep learning for spectral clustering was introduced by \citet{Wada2019Spectral}. Their method starts by identifying hub points, which serve as the core of clusters. These hub points are then passed to a deep network to obtain the cluster labels for the remaining points.

\subsection{Graph similarity metrics}
Every graph clustering method needs a metric to construct pairwise similarities. A shared neighbor similarity was introduced by \citet{ZHANG2021Spectral}. They applied their method to \textit{attributed graphs}, a special type of graph where each node has feature attributes. They used shared neighbor similarities to highlight clusters’ discontinuity. The concept of shared neighbors could be traced back to Jarvis–Patrick algorithm \cite{Jarvis1973Clustering}. It is important to mention the higher cost associated with shared neighbor similarity. Because all neighbors have to be matched, instead of computing one value such as the Euclidean distance. 

Another way of constructing pairwise similarities was introduced by \citet{Wen2020Spectral}, where they utilized Locality Preserving Projection (LPP) and hypergraphs. First, all points are projected onto a space with reduced dimensionality. The pairwise similarities are constructed using a heat kernel (Equation \ref{Eq-heatkernel}). Second, a hypergraph Laplacian matrix $L_H$ is used to replace the regular graph Laplacian matrix $L$. Hypergraphs would help to capture the higher relations between vertices. Two things needed to be considered when applying this method: 1) the $\sigma$ parameter in the heat kernel needs careful tuning \cite{Zelnik2005Self}, and 2) the computational cost for hypergraph Laplacian matrix $L_H$. Density information were incorporated into pairwise similarity construction by \citet{Kim2021KNN}. The method defines (locally dense points) that are separated from each other by (locally sparse points). This approach falls under the category of DBSCAN clustering \cite{Ester1996Density}. These methods are iterative by nature and need a stopping criterion to be defined.

\begin{equation}
	A_{i,j} = exp^{\frac{\left\Vert x_i - x_j \right\Vert^2_2}{2\sigma ^2}}
	\label{Eq-heatkernel}
\end{equation}

Considering the literature on graph representation learning, it is evident that SpectralNet \cite{shaham2018spectralnet}: 1) offers a cost-efficient method to perform graph clustering using deep networks and 2) it does not require labeled datasets. The problem is that it uses $k$-nearest neighbor graph with distance metric. This restricts points from pairing with more neighbors if they are in a close proximity. A suitable alternative would be a similarity metric based on random projection trees \cite{Dasgupta2008Random,Freund2008Learning}. rpTrees similarity were already used in spectral clustering by \cite{Yan2009Fast,Yan2019Similarity}. But they are yet to be extended to graph clustering using deep networks.

\section{SpectralNet and  pairwise similarities}
\label{ProposedApproach}
\begin{figure}[h]
	\centering
	\includegraphics[width=\textwidth,height=20cm,keepaspectratio]{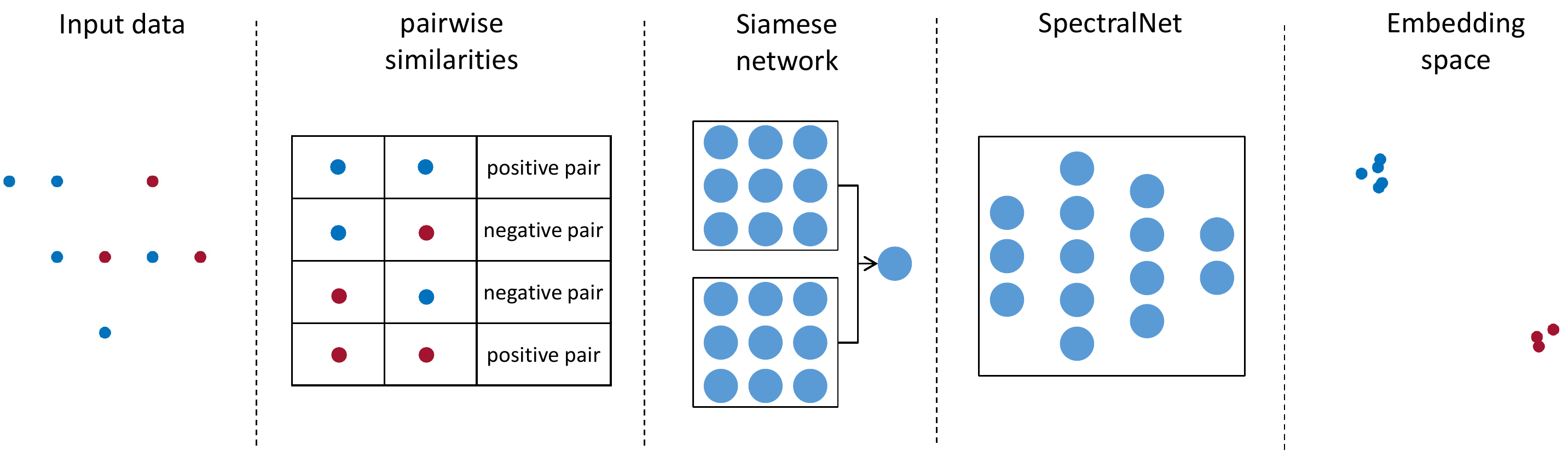}	
	\caption{An outline of the used algorithm. (Best viewed in color)}
	\label{Fig:outline}
\end{figure}

The proposed rpTree similarity metric was used in SpectralNet alongside the distance metric that was used for $k$-nearest neighbor graph. The SpectralNet algorithm consists of four steps: 1) identifying positive and negative pairs, 2) running Siamese net using positive and negative pairs to construct the affinity matrix $A$, 3) SpectralNet that maps points onto an embedding space, and 4) clusters are detected by running $k$-means in the embedding space. An illustration of these steps is shown in Fig.\ \ref{Fig:outline}. The next subsection explains the used neural networks (Siamese and SpectralNet). The discussion of similarity metrics and their complexity is introduced in the following subsections.

\subsection{SpectralNet}
\label{SpectralNet}
The first step in SpectralNet is the Siamese network, which consists of two or more neural networks with the same structure and parameters. These networks has a single output unit that is connected to the output layers of the networks in the Siamese net. For simplicity let us assume that the Siamese net consists of two neural networks $N_1$ and $N_2$. Both networks received inputs $x_1$ and $x_2$ respectively, and produce two outputs $z_1$ and $z_2$. The output unit compared the two outputs using the Euclidean distance. The distance should be small if $x_1$ and $x_2$ are a positive pair, and large if they are a negative pair. The Siamese net is trained to minimize contrastive loss, that is defined as:
\begin{equation}
	L_{\text{contrastive}}=\begin{cases}
		\left\Vert z_1 - z_2 \right\Vert^2, & \text{if $(x_1,x_2)$ is a positive pair}\\
		max(c-\left\Vert z_1 - z_2 \right\Vert,0), & \text{if $(x_1,x_2)$ is a negative pair},
	\end{cases}
	\label{Eq-Siamese}
\end{equation}
\noindent
where $c$ is a constant that is usually set to 1. Then the Euclidean distance obtained via the Siamese net $\left\Vert z_1 - z_2 \right\Vert$ is used in the heat kernel (see Equation \ref{Eq-heatkernel}) to find the similarities between data points and construct the affinity matrix $A$.

The SpectralNet uses a gradient step to optimize the loss function $L_{SpectralNet}$:

\begin{equation}
	L_{SpectralNet} = \frac{1}{m^2} \sum_{i,j=1}^{m} a_{i,j} \left\Vert y_i - y_j \right\Vert^2
	\label{Eq-SpectralNet}
\end{equation}
\noindent
where $m$ is the batch size; $a$ of size $m \times m$ is the affinity matrix of the sampled points; $y_i$ and $y_j$ are the expected labels of the samples $x_i$ and $x_j$. But the optimization of this functions is constrained, since the last layer is set to be a constraint layer that enforces orthogonalization. Therefore, SpectralNet has to alternate between orthogonalization and gradient steps. Each of these steps uses a different random batch $m$ from the original data $X$. Once the SpectralNet is trained, all samples $x_1, x_2, \cdots, x_n$ are passed through network to get the predictions $y_1, y_2, \cdots, y_n$. These predictions represent coordinates on the embedding space, where $k$-means operates and finds the clustering.

\subsection{Constructing pairwise similarities using $k$-nn}
\label{Constructing-pairs-knn}
\begin{figure}[h]
	\begin{minipage}[b]{0.6\linewidth}
		\centering
		\includegraphics[width=0.99\textwidth,height=20cm,keepaspectratio]{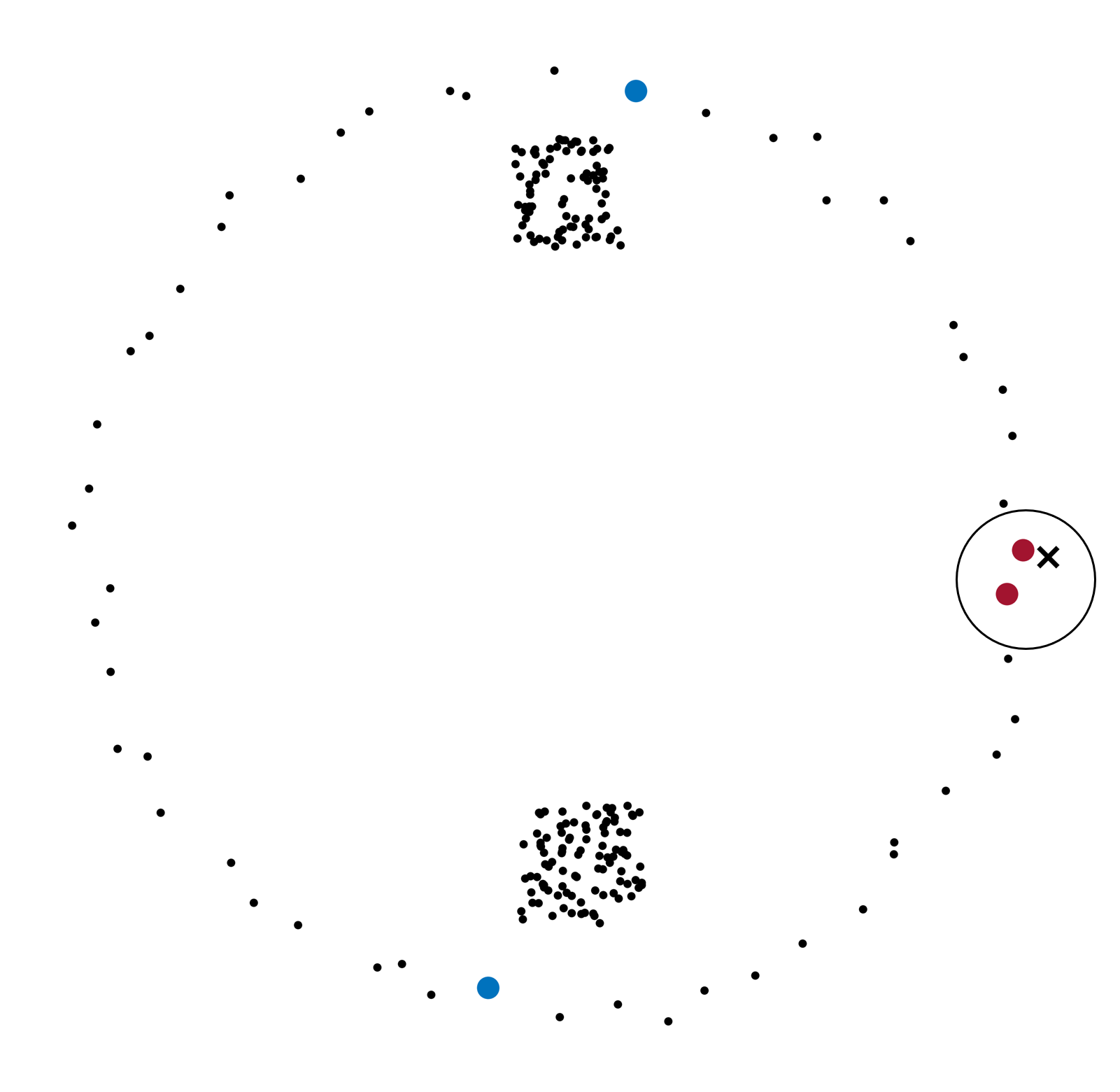}
	\end{minipage}
	\begin{minipage}[b]{0.4\linewidth}
		\centering
		\includegraphics[width=0.9\textwidth,height=20cm,keepaspectratio]{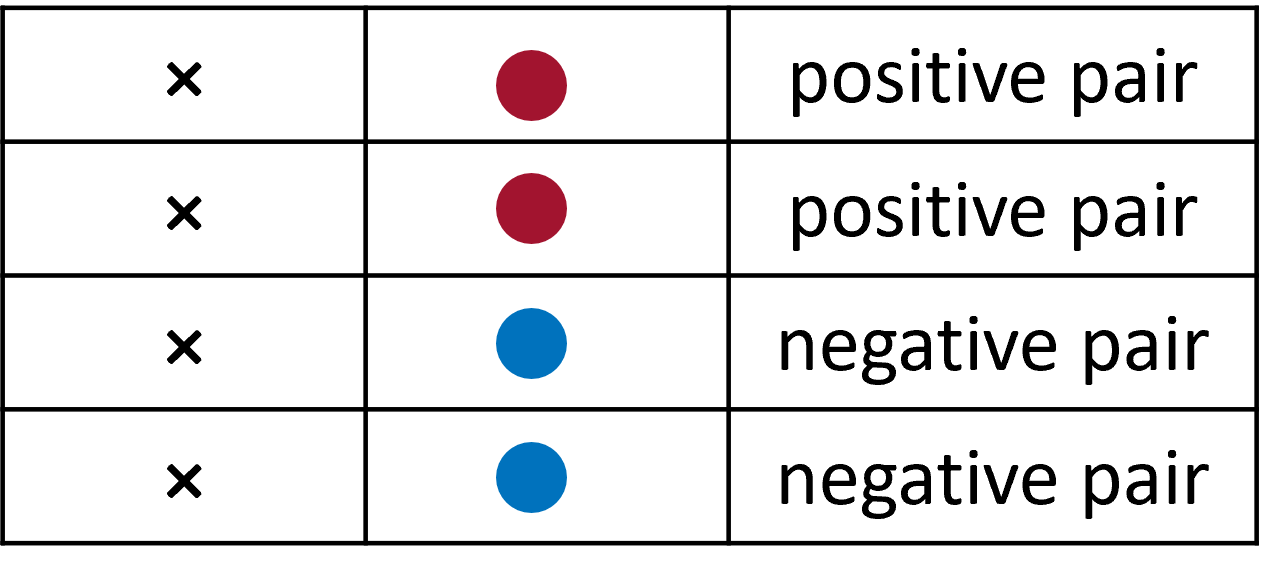}
		\vspace{14ex}
	\end{minipage}
	\caption{Constructing positive and negative pairs using $k$-nn search; red points are the nearest neighbors when $k=2$; blue points are selected randomly (Best viewed in color)}
	\label{Fig:knn-similarity}
\end{figure}

The original algorithm of SpectralNet \cite{shaham2018spectralnet} has used $k$-nearest neighbor graph with distance metric to find positive and negative pairs. The positive pairs are the nearest neighbors according to the selected value of $k$, the original method has $k$ set to be $2$. The negative pairs were selected randomly from the farther neighbors. An illustration of this process is shown in Fig.\ \ref{Fig:knn-similarity}

Restricting the points to have a fixed number of positive pairs can be a disadvantage of using $k$-nn. That is a problem we are trying to overcome by using rpTrees to construct positive and negative pairs. In rpTrees, there is no restriction on how many number of pairs for individual points. It depends on how many points ended up in the same leaf node.

\subsection{Constructing pairwise similarities using rpTrees}
\label{Constructing-pairs-rpTree}
rpTrees start by choosing a random direction $\overrightarrow{r}$ from the unit sphere $S^{D-1}$, where $D$ is the number of dimensions. All points in the current node $W$ are projected onto $\overrightarrow{r}$. On that reduced space $\mathbb{R}^{D-1}$, the algorithm picks a dimension uniformly at random and chooses the split point $c$ randomly between $\lbrack \frac{1}{4}, \frac{3}{4} \rbrack$. The points less than the split point $x < c$ are placed in the left child $W_L$, and the points larger than the split point $x > c$ are placed in the right child $W_R$. The algorithm continues to partition the points recursively, and stops when the split produces a node with points less than the leaf size parameter $n_0$.

\begin{figure}[h]
	\begin{minipage}[b]{0.7\linewidth}
		\centering
		\includegraphics[width=0.99\textwidth,height=20cm,keepaspectratio]{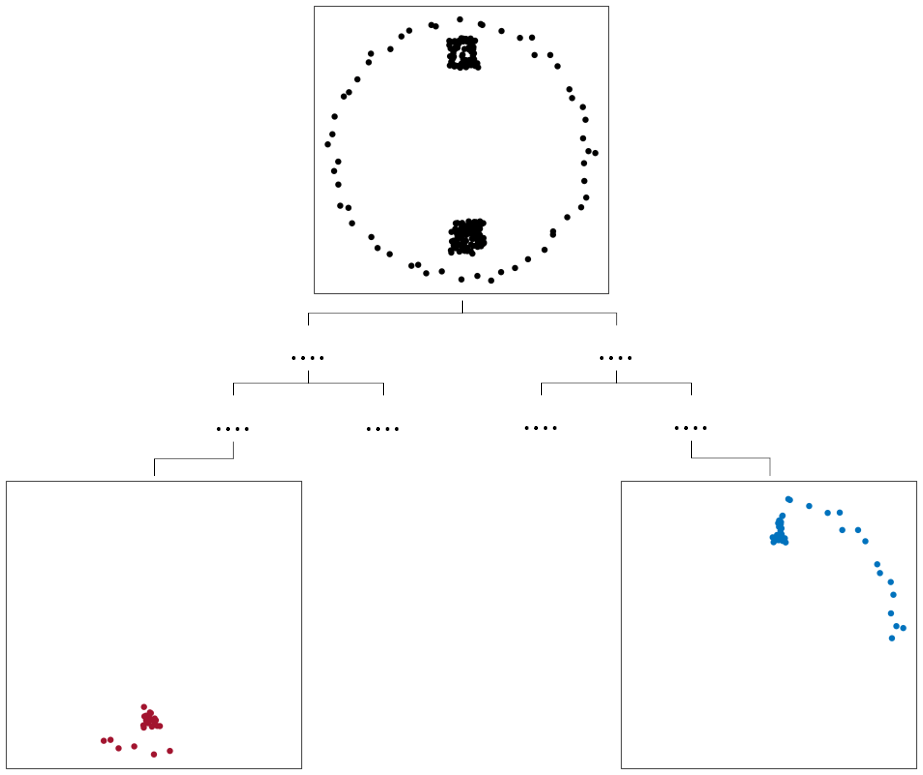}
	\end{minipage}	
	\begin{minipage}[b]{0.28\linewidth}
		\centering
		\includegraphics[width=0.99\textwidth,height=20cm,keepaspectratio]{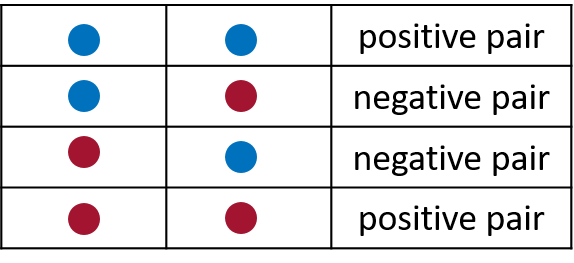}
		\vspace{15ex}
	\end{minipage}
	\caption{Constructing positive and negative pairs using rpTree. (Best viewed in color)}
	\label{Fig:rpTree-similarity}
\end{figure}

To create positive pairs for the Simese net, we pair all points in one leaf node. So, points that fall onto the same leaf node are considered similar, and we mark them as positive pairs. For negative pairs, we pick one leaf node $W_x$, and from the remaining set of leaf nodes we randomly pick $W_y$. Then, we pair all points in $W_x$ with the points in $W_y$, and mark them as negative pairs (Equation \ref{Eq-pn-pairs}). An illustration of this process is shown in Fig.\ \ref{Fig:rpTree-similarity}.

\begin{equation}
	\begin{split}
		&(p,q) \in E(positive) \Leftrightarrow p \in W_x\ and\  q \in W_x\\
		&(p,q) \in E(negative) \Leftrightarrow p \in W_x\ and\  q \in W_y.
	\end{split}
	\label{Eq-pn-pairs}
\end{equation}

\subsection{Complexity analysis for computing pairwise similarities}
\label{Complexityanalysis}
We will use the number of positive pairs to analyze the complexity of the similarity metric used in the original SpectralNet method and the metric proposed in this paper. The original method uses the nearest $k$ neighbors as positive pairs. This is obviously grows linearly with $n$, since we have to construct $n \times k$ pairs and pass them to the Siamese net \cite{Bromley1993Signature}.

Before we analyze the proposed metric, we have to find how many points will fall into a leaf node of an rpTree. This question is usually asked in proximity graphs \cite{Gilbert1961Random}. If we place a squared tessellation $T$ on top of $n$ data points (please refer to section 9.4.1 by Barthelemy \cite{barthelemy2017morphogenesis} for more elaboration). $T$ has an area of $n$ and a side length of $\sqrt{n}$. Each small square $s$ in $T$ has an area of $\log(n)$. The probability of having more than $k$ neighbors in $s$ is $P(l>k)$, where $l = k+1,\cdots,n$. The probability $P(l>k)$ follows the homogeneous Poisson process. This probability approximately equals $\frac{1}{n}$, which is very small, suggesting there is a significant chance of having at most $\log(n)$ neighbors in a square $s$. Since rpTrees follow the same approach of partitioning the search space, it is safe to assume that each leaf node would have at most $\log(n)$ data points.

The proposed metric depends on the number of leaf nodes in the rpTree and the number of points in each leaf node ($n_0$). The leaf size $n_0$ is a parameter that can be tuned before running rpTree. It also determines how many leaf nodes in the tree. Because the method will stop partitioning when the number of points in a leaf node reaches a minimum limit. Then we have to pair all the points in the leaf node.

\begin{figure}[h]
	\centering
	\includegraphics[width=0.8\textwidth,height=20cm,keepaspectratio]{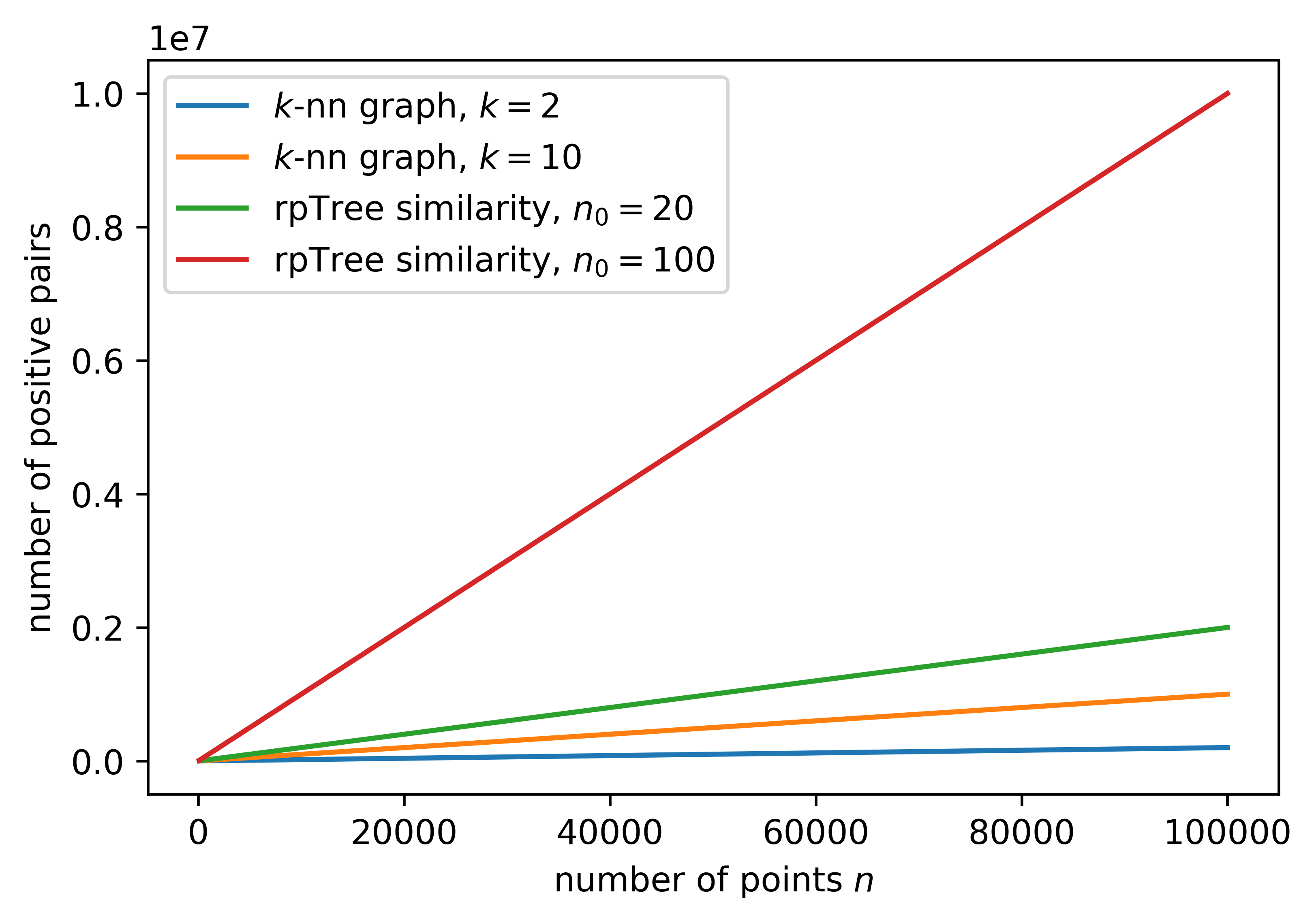}	
	\caption{The expected number of positive pairs using $k$-nn and rpTree similarities. (Best viewed in color)}
	\label{Fig:similarity-plot}
\end{figure}

To visualize this effect, we have to fix all parameters and vary $n$. In Fig.\ \ref{Fig:similarity-plot}, we set $k$ to be $2$ and $10$. The leaf size $n_0$ was set to $20$ and $100$. The number of points $n$ was in the interval $[100,100000]$. With $k$-nn graph we need $n \times k$ positive pairs, and in rpTree similarity we need ${n_0}^2 \times \frac{n}{n_0} = n \times n_0$ positive pairs. So, both similarity metrics grow linearly with $n$. The main difference is how the points are partitioned. $k$-nn graph uses $kd$-tree which produces the same partition with each run making the number of positive pairs fixed. But rpTrees partitions points randomly, so the number of positive pairs will deviate from $n \times n_0$.

\begin{figure}[h]
	\centering
	\includegraphics[width=0.24\textwidth]{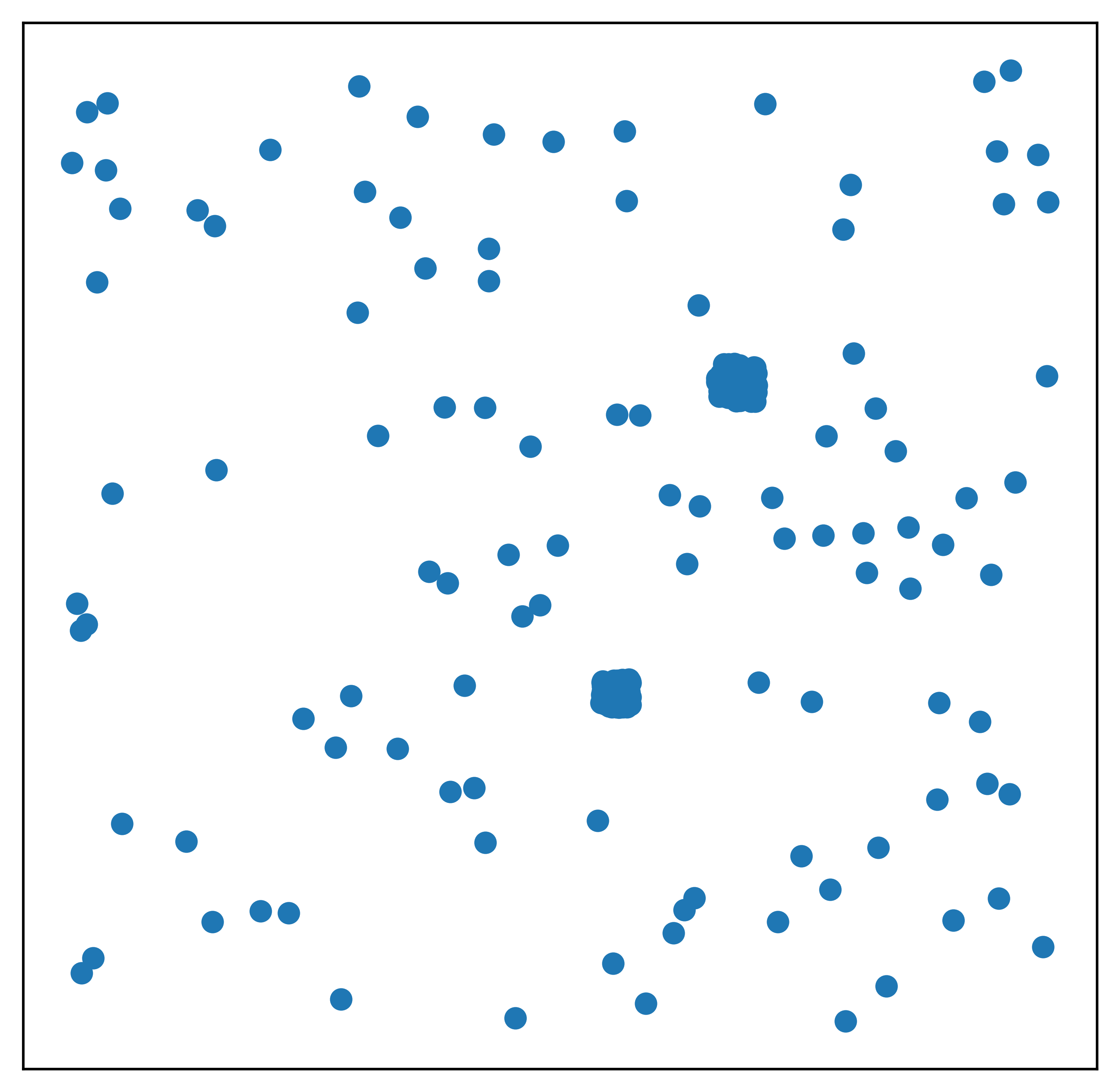}  
	\includegraphics[width=0.24\textwidth]{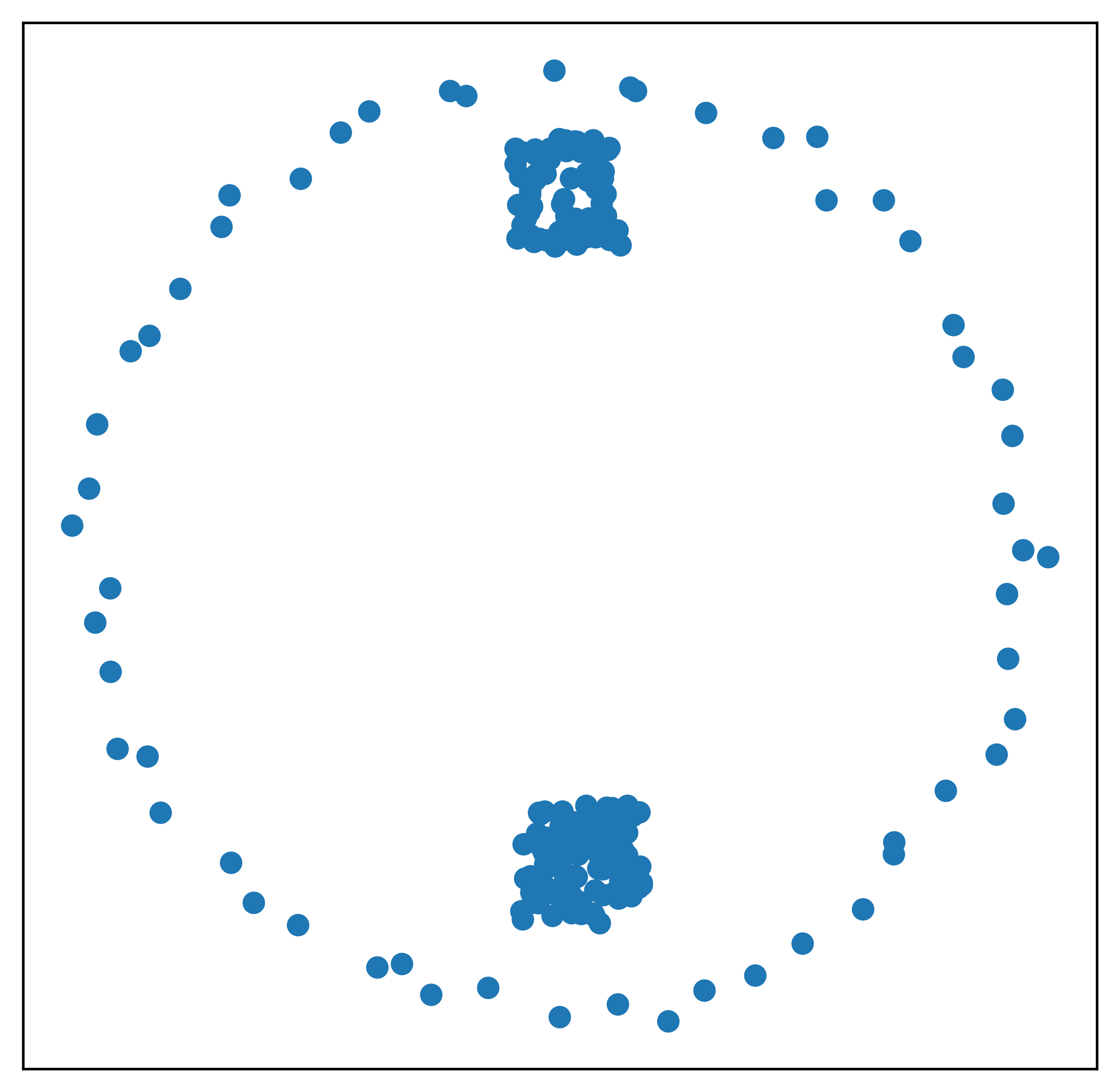}
	\includegraphics[width=0.24\textwidth]{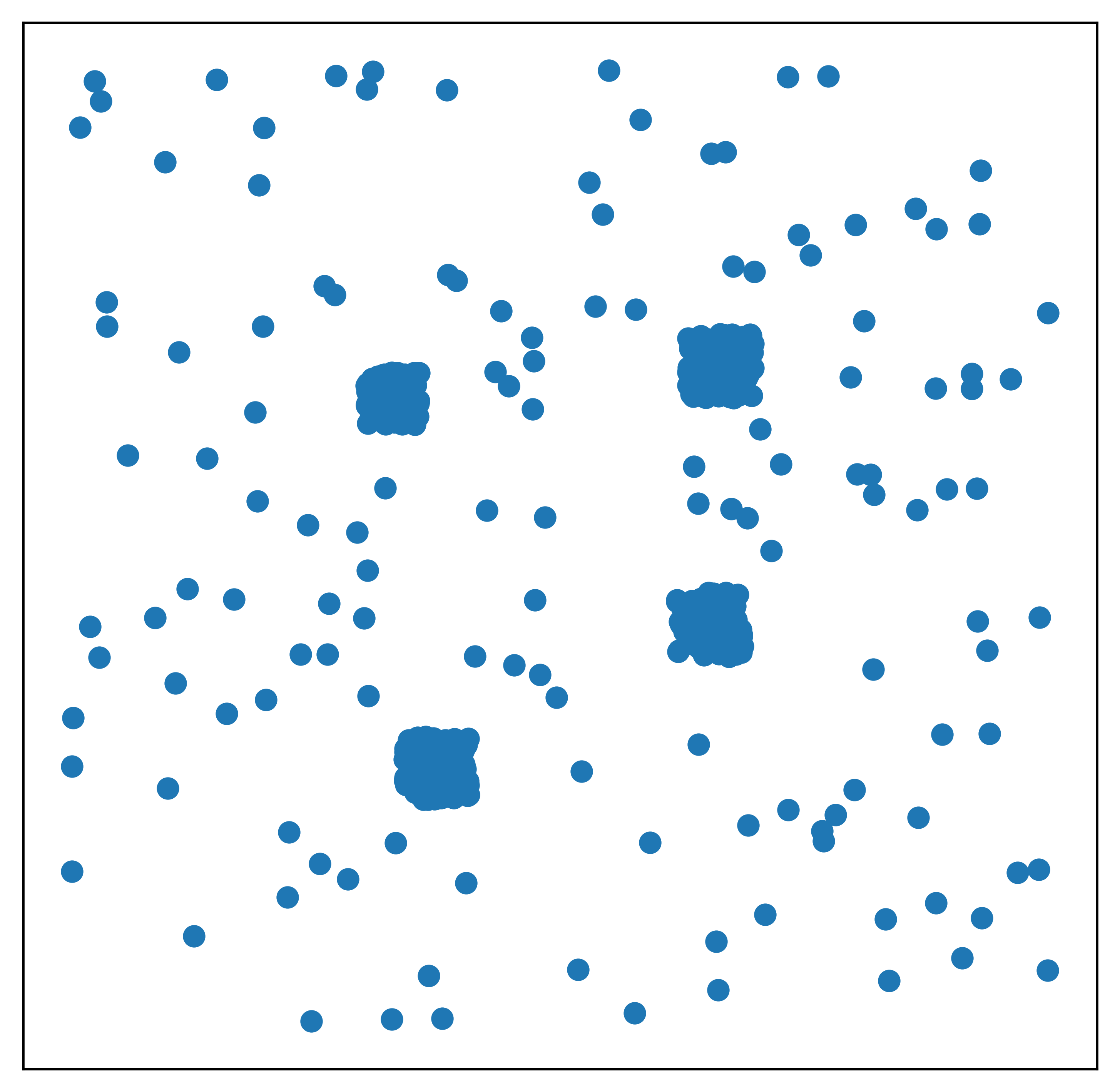}  
	\includegraphics[width=0.24\textwidth]{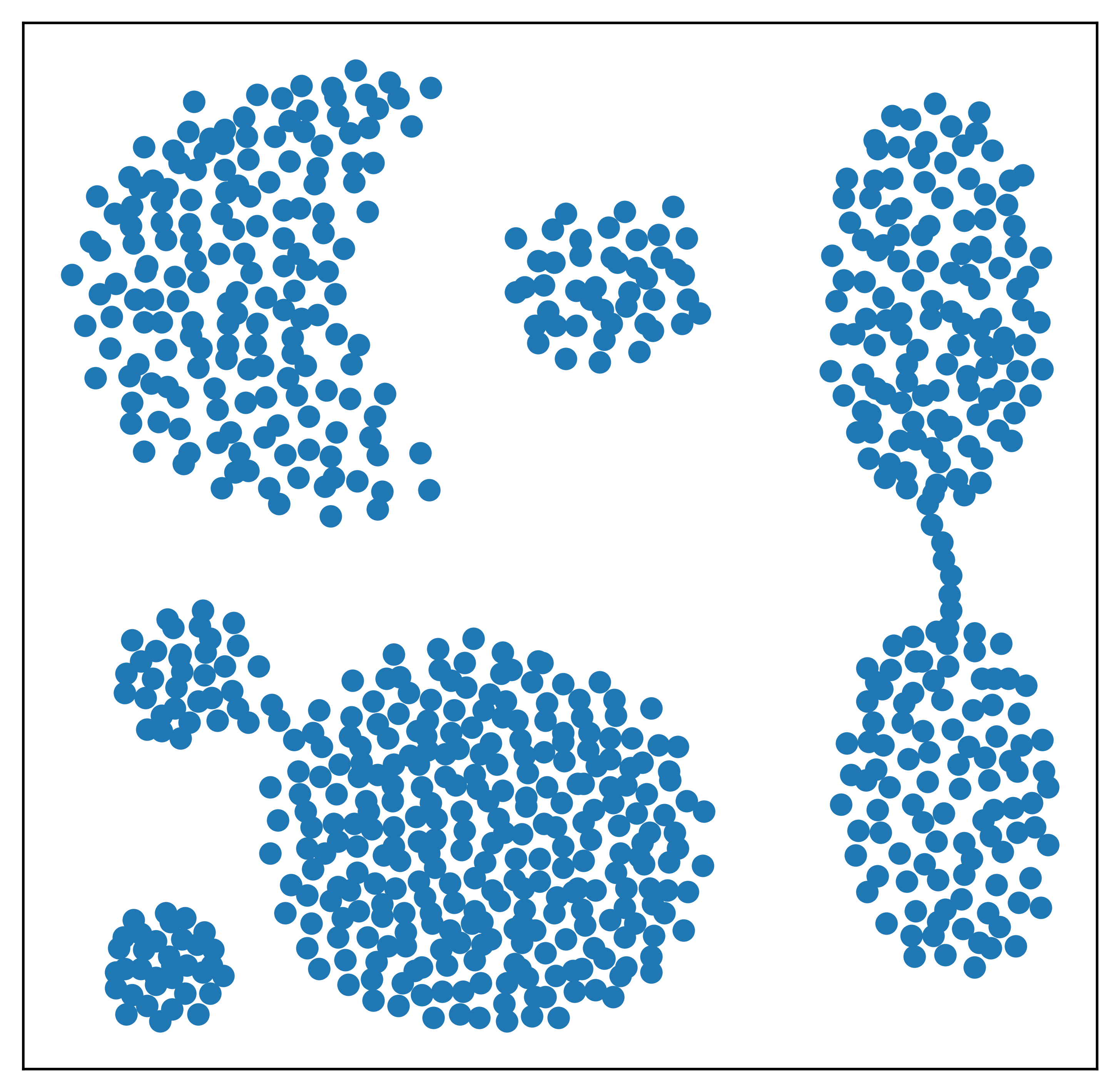}	  
	\caption{Synthetic datasets used in the experiments; from left to right \texttt{Dataset 1} to \texttt{Dataset 4}.}
	\label{Fig:Datasets}
\end{figure}

\section{Experiments and discussions}
\label{Experiments}
In our experiments we compared the similartity metrics using $k$-nearest neighbor and rpTree, in terms of: 1) clustering accuracy and 2) storage efficiency. The clustering accuracy was measured using Adjusted Rand Index (\texttt{ARI}) \cite{Hubert1985Comparing}. Given the true grouping $T$ and the predicted grouping $L$, \texttt{ARI} is computed using pairwise comparisons. $n_{11}$ if the pair belong to the same cluster in $T$ and $L$ groupings, and $n_{00}$ if the pair in different clusters in $T$ and $L$ groupings. $n_{01}$ and $n_{10}$ if there is a mismatch between $T$ and $L$. \texttt{ARI} is defined as:

\begin{equation}
	ARI(T,L)=\frac{2(n_{00}n_{11}-n_{01}n_{10})}{(n_{00}+n_{01})(n_{01}+n_{11})+(n_{00}+n_{10})(n_{10}+n_{11})}\ .
	\label{Eq-ARI}
\end{equation}
\noindent
The storage efficiency was measured by the number of total pairs used. We avoid using machine dependent metrics like the running time.

We also run additional experiments to investigate how the rpTrees parameters are affecting the similarity metric based on rpTree. The first parameter was the leaf size parameter $n_0$, which determines the minimum number of points in a leaf node. The second parameter was how to select the projections direction. There are a number of methods to choose the random direction. We tested these methods to see how they would affect the performance.

The two dimensional datasets used in our experiments are shown in Fig.\ \ref{Fig:Datasets}. The remaining datasets were retrieved from scikit-learn library \cite{scikit-learn, sklearn_api}, except for the \texttt{mGamma} dataset which was downloaded from UCI machine learning repository \cite{Dua2019UCI}. All experiments were coded in python 3 and run on a machine with 20 GB of memory and
a 3.10 GHz Intel Core i5-10500 CPU. The code can be found on \url{https://github.com/mashaan14/RPTree}.

\begin{figure}[h]
	\centering
	\includegraphics[width=\textwidth]{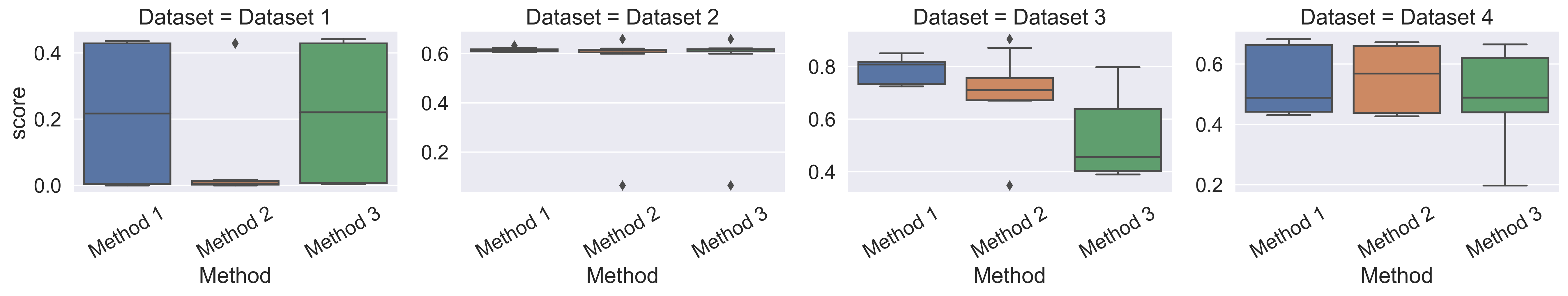}  
	\includegraphics[width=\textwidth]{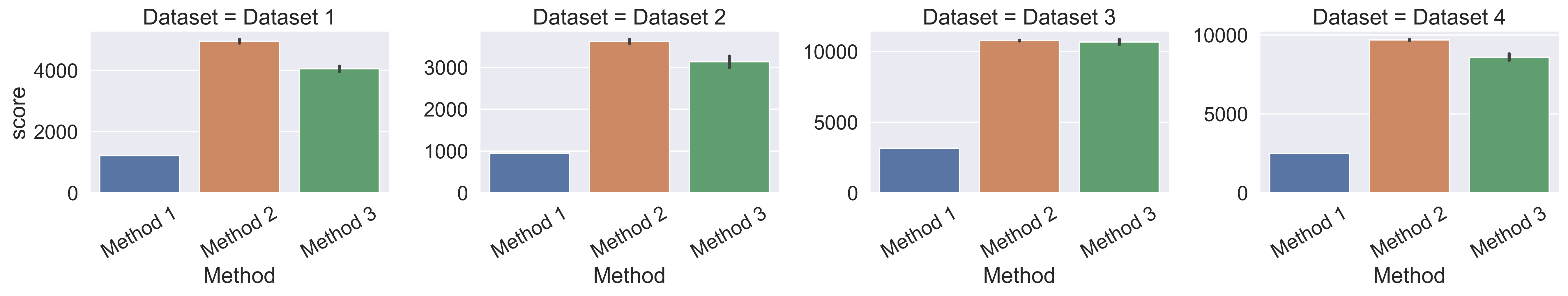}
	\caption{Experiments with synthetic datasets; \texttt{Method 1} is $k$-nn graph with $k=2$, \texttt{Method 2} is $k$-nn graph with varying $k$, and \texttt{Method 3} is rpTree similarity with $n_0=20$; (top) ARI scores for 10 runs, (bottom) number of total pairs. (Best viewed in color)}
	\label{Fig:Results-synthetic-01}
\end{figure}

\begin{figure}[h]
	\centering
	\includegraphics[width=\textwidth]{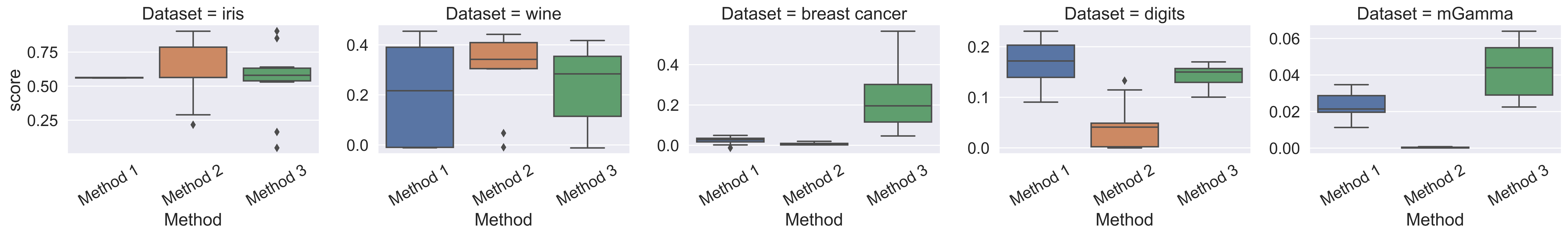}  
	\includegraphics[width=\textwidth]{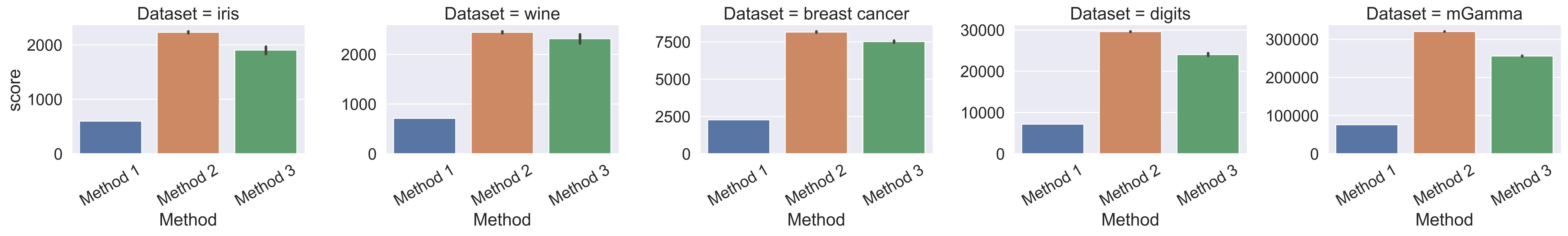}
	\caption{Experiments with real datasets; \texttt{Method 1} is $k$-nn graph with $k=2$, \texttt{Method 2} is $k$-nn graph with varying $k$, and \texttt{Method 3} is rpTree similarity with $n_0=20$; (top) ARI scores for 10 runs, (bottom) number of total pairs. (Best viewed in color)}
	\label{Fig:Results-real-01}
\end{figure}

\subsection{Experiments using $k$-nn and rpTree similarity metrics}
\label{k-nn-and-rpTree}
Three methods were used in this experiment. \texttt{Method 1} is the original SpectralNet method by Shaham et al \cite{shaham2018spectralnet}. \texttt{Method 2} was developed by Alshammari et al. \cite{Alshammari2021Refining}, it sets $k$ dynamically based on the statistics around the points. \texttt{Method 3} is the proposed method which uses an rpTree similarity instead of $k$-nn graph.

With the four synthetic datasets, all three methods delivered similar performances shown in Fig.\ \ref{Fig:Results-synthetic-01}. Apart from \texttt{Dataset 3}, where rpTree similarity performed lower than other methods. This could be attributed to how the clusters are distributed in this dataset. The rpTree splits separated points from the same cluster, which lowers the \texttt{ARI}. \texttt{Method 2} has the maximum number of pairs over all three methods. The number of pairs in \texttt{Method 2} and \texttt{Method 3} deviated slightly from the mean, unlike \texttt{Method 1} which has the same number of pairs with each run because $k$ was fixed ($k=2$).

rpTrees similarity outperformed other methods in three out of the five real datasets \texttt{iris}, \texttt{breast cancer}, and \texttt{mGamma} as shown in Fig.\ \ref{Fig:Results-real-01}. $k$-nn with Euclidean distance performed poorly in \texttt{breast cancer}, which suggests that connecting to two neighbors was not enough to accurately detect the clusters. Yan et al. reported a similar finding where clustering using rpTree similarity was better than clustering using Gaussian kernel with Euclidean distance \cite{Yan2019Similarity}. They showed the heatmap of the similarity matrix generated by the Gaussian kernel and by rpTree.

As for the number of pairs, the proposed similarity metric was the second lowest method that used total pairs across all five datasets. Because of the randomness involved in rpTree splits, the proposed similarity metric has a higher standard deviation for the number of total pairs.

\begin{figure}[h]
	\centering
	\includegraphics[width=\textwidth]{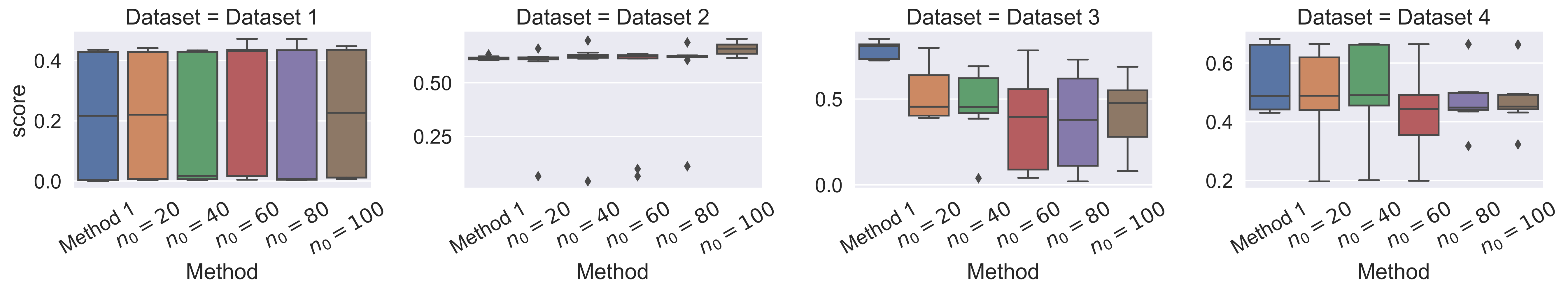}  
	\includegraphics[width=\textwidth]{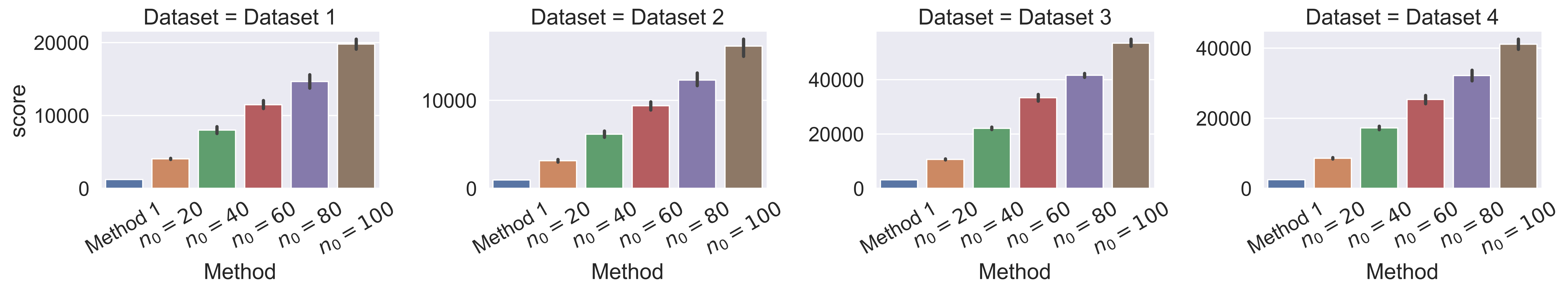}
	\caption{Experiments with synthetic datasets; \texttt{Method 1} is $k$-nn graph with $k=2$, other methods use rpTree similarity with varying $n_0$; (top) ARI scores for 10 runs, (bottom) number of total pairs. (Best viewed in color)}
	\label{Fig:Results-synthetic-02}
\end{figure}

\begin{figure}[h]
	\centering
	\includegraphics[width=\textwidth]{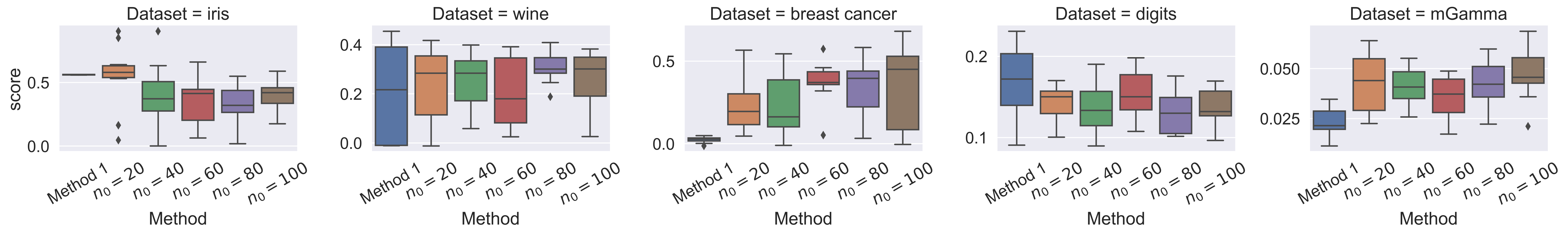}  
	\includegraphics[width=\textwidth]{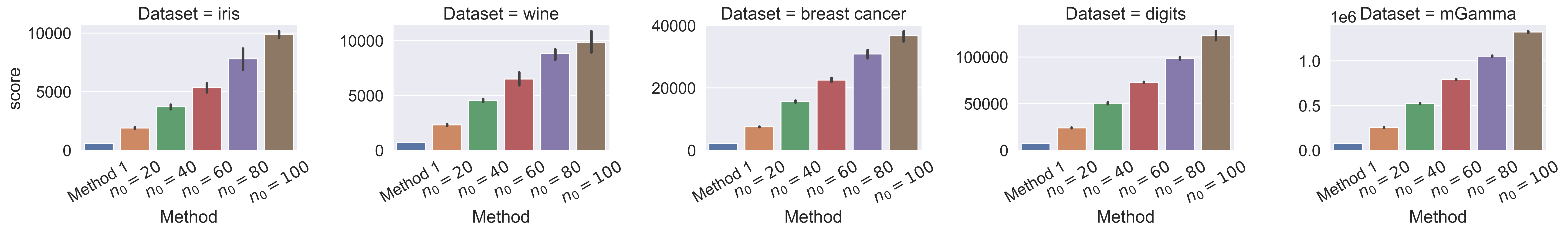}
	\caption{Experiments with real datasets; \texttt{Method 1} is $k$-nn graph with $k=2$, other methods use rpTree similarity with varying $n_0$; (top) ARI scores for 10 runs, (bottom) number of total pairs. (Best viewed in color)}
	\label{Fig:Results-real-02}
\end{figure}

\subsection{Investigating the influence of the leaf size parameter $n_0$}
\label{Investigating-w}

One of the important parameters in rpTrees is the leaf size $n_0$. It determines when the rpTree stops growing. If the number of points in a leaf node is less than the leaf size $n_0$, that leaf node would not be split further.

By looking at the clustering performance in synthetic datasets shown in Fig.\ \ref{Fig:Results-synthetic-02} (top), we can see that we are not gaining much by increasing the leaf size $n_0$. In fact, increasing the leaf size $n_0$ might affect the clustering accuracy like what happened in \texttt{Dataset 3}. The number of pairs is also related with the leaf size $n_0$, as it grows with $n_0$. This is shown in Fig.\ \ref{Fig:Results-synthetic-02} (bottom).

Increasing the leaf size $n_0$ helped us to get higher \texttt{ARI} with \texttt{breast cancer} and \texttt{mGamma} as shown in Fig.\ \ref{Fig:Results-real-02}. With other real datasets it was not improving the clustering accuracy measured by \texttt{ARI}. We also observed that the number of pairs increases as we increase $n_0$.

Ram and Sinha \cite{Ram2019Revisiting} provided a discussion on how to set the parameter $n_0$. They stated that $n_0$ controls the balance between global search and local search. Overall, they stated that $n_0$ effect on search accuracy is ``quite benign’’.

\subsection{Investigating the influence of the dispersion of points along the projection direction}
\label{Investigating-r}

The original algorithm of rpTrees \cite{Dasgupta2008Random} suggests using a random direction selected at random. But a recent application of rpTree by \citet{Yan2021Nearest} recommended picking three random directions ($nTry=3$) and use the one that provides the maximum spread of data points. To investigate the effect of this parameter, we used four methods for picking a projection direction: 1) picking one random direction, 2) picking three random direction ($nTry=3$) and use the one with maximum spread, 3) picking nine random directions ($nTry=9$), and 4) using principal component analysis (PCA) to find the direction with the maximum spread.

\begin{figure}[h]
	\centering
	\includegraphics[width=\textwidth]{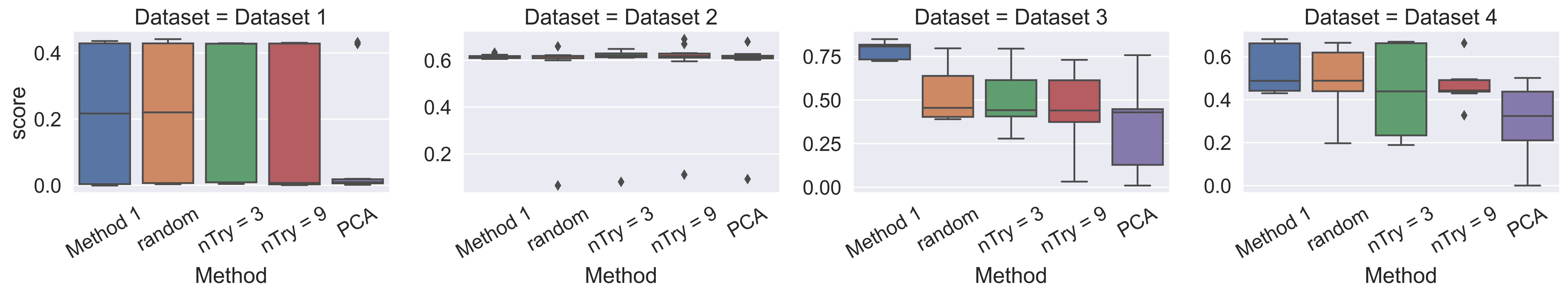}  
	\includegraphics[width=\textwidth]{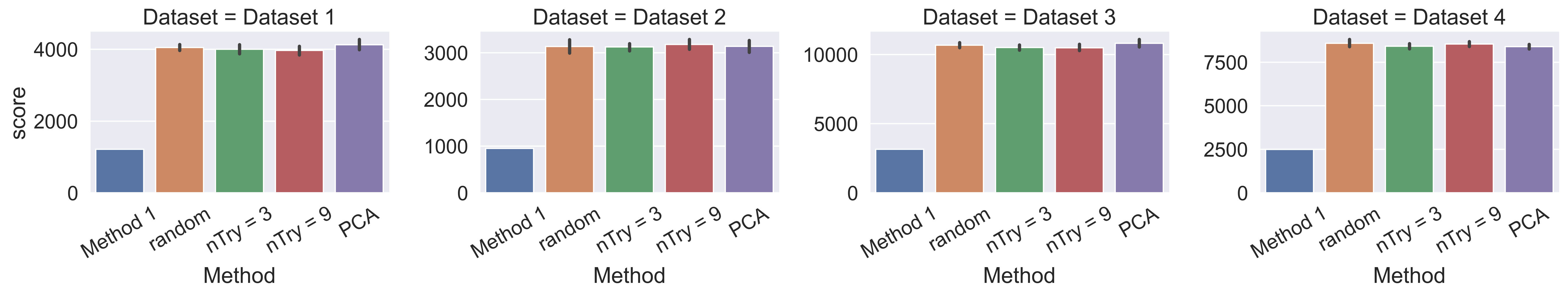}
	\caption{Experiments with synthetic datasets; \texttt{Method 1} is $k$-nn graph with $k=2$, other methods use different strategies to pick the projection direction; (top) ARI scores for 10 runs, (bottom) number of total pairs. (Best viewed in color)}
	\label{Fig:Results-synthetic-03}
\end{figure}

\begin{figure}[h]
	\centering
	\includegraphics[width=\textwidth]{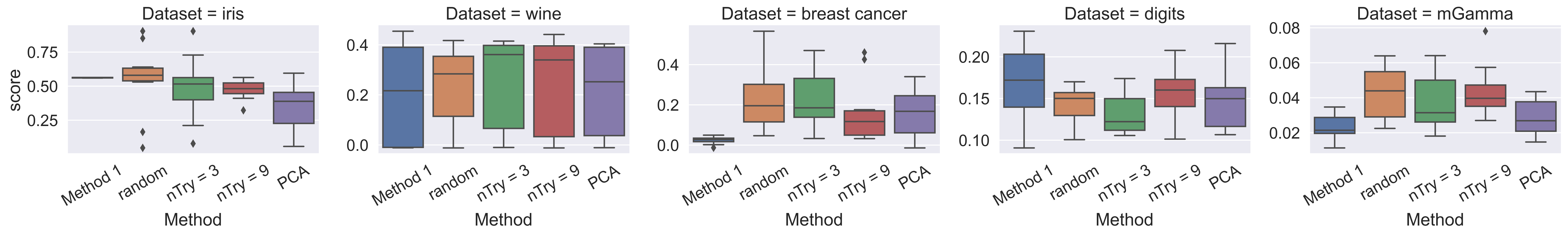}
	\includegraphics[width=\textwidth]{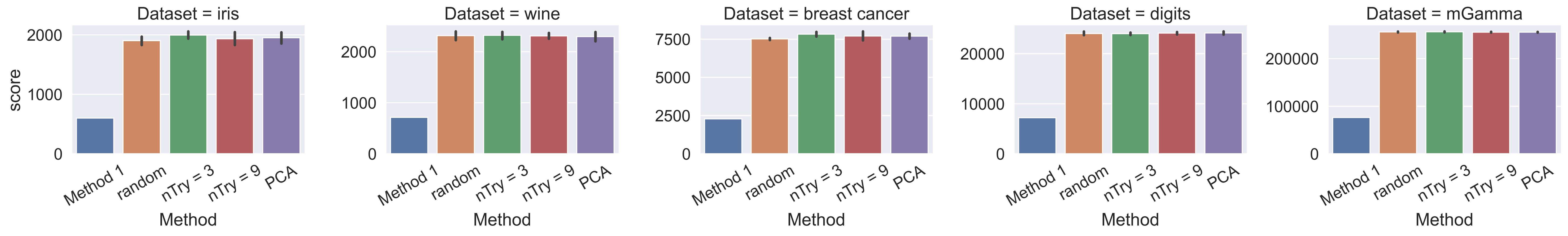}
	\caption{Experiments with real datasets; \texttt{Method 1} is $k$-nn graph with $k=2$, other methods use different strategies to pick the projection direction; (top) ARI scores for 10 runs, (bottom) number of total pairs. (Best viewed in color)}
	\label{Fig:Results-real-03}
\end{figure}

By looking at \texttt{ARI} numbers for synthetic datasets (Fig.\ \ref{Fig:Results-synthetic-03}) and real datasets (Fig.\ \ref{Fig:Results-real-03}), we observed that we are not gaining much by trying to maximize the spread of projected points. This parameter has very little effect. Also, all methods with different strategies to pick the projection direction have used the same number of pairs.

\begin{figure}[h]
	\centering
	\includegraphics[width=0.4\textwidth]{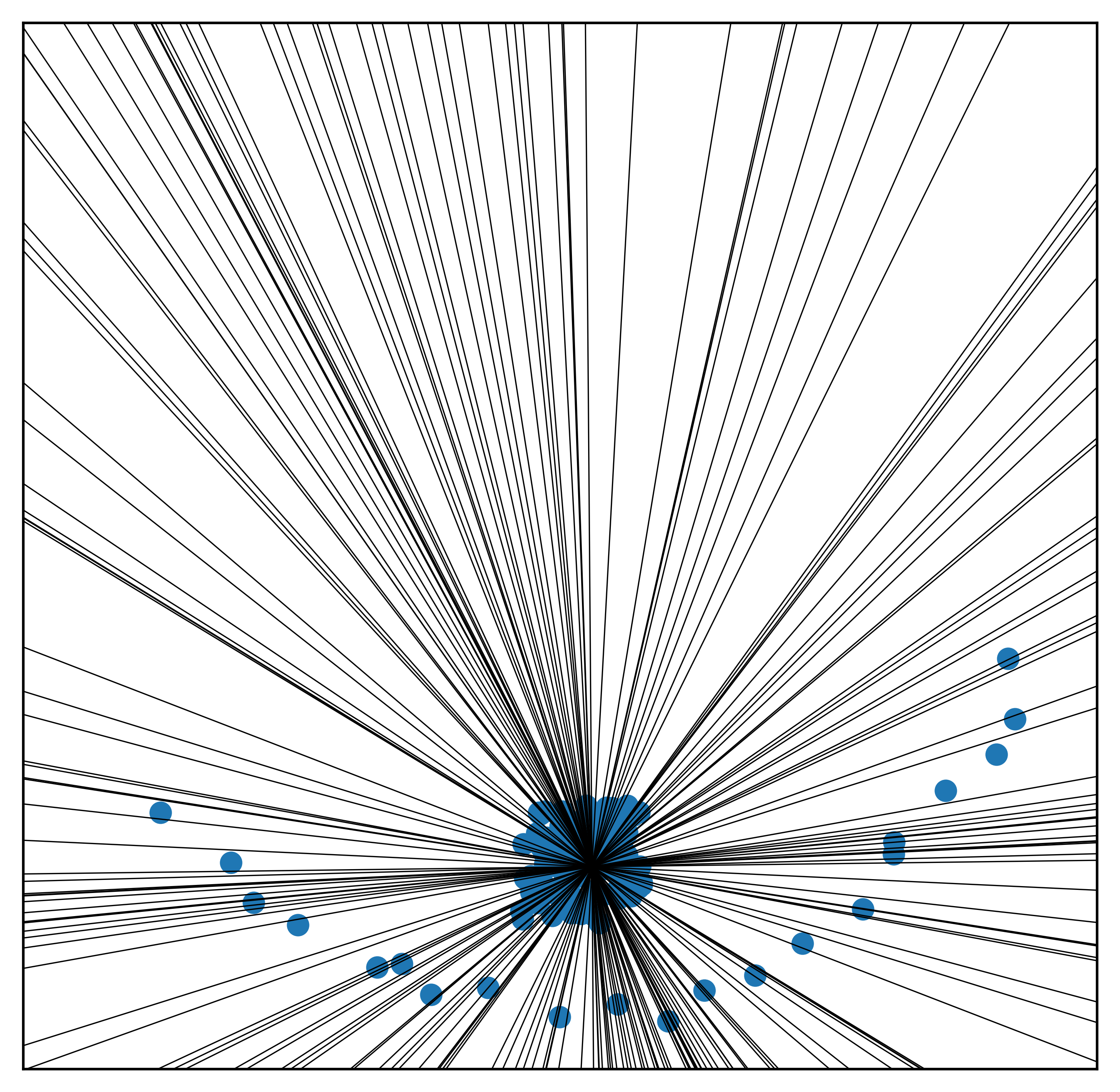}  
	\includegraphics[width=0.4\textwidth]{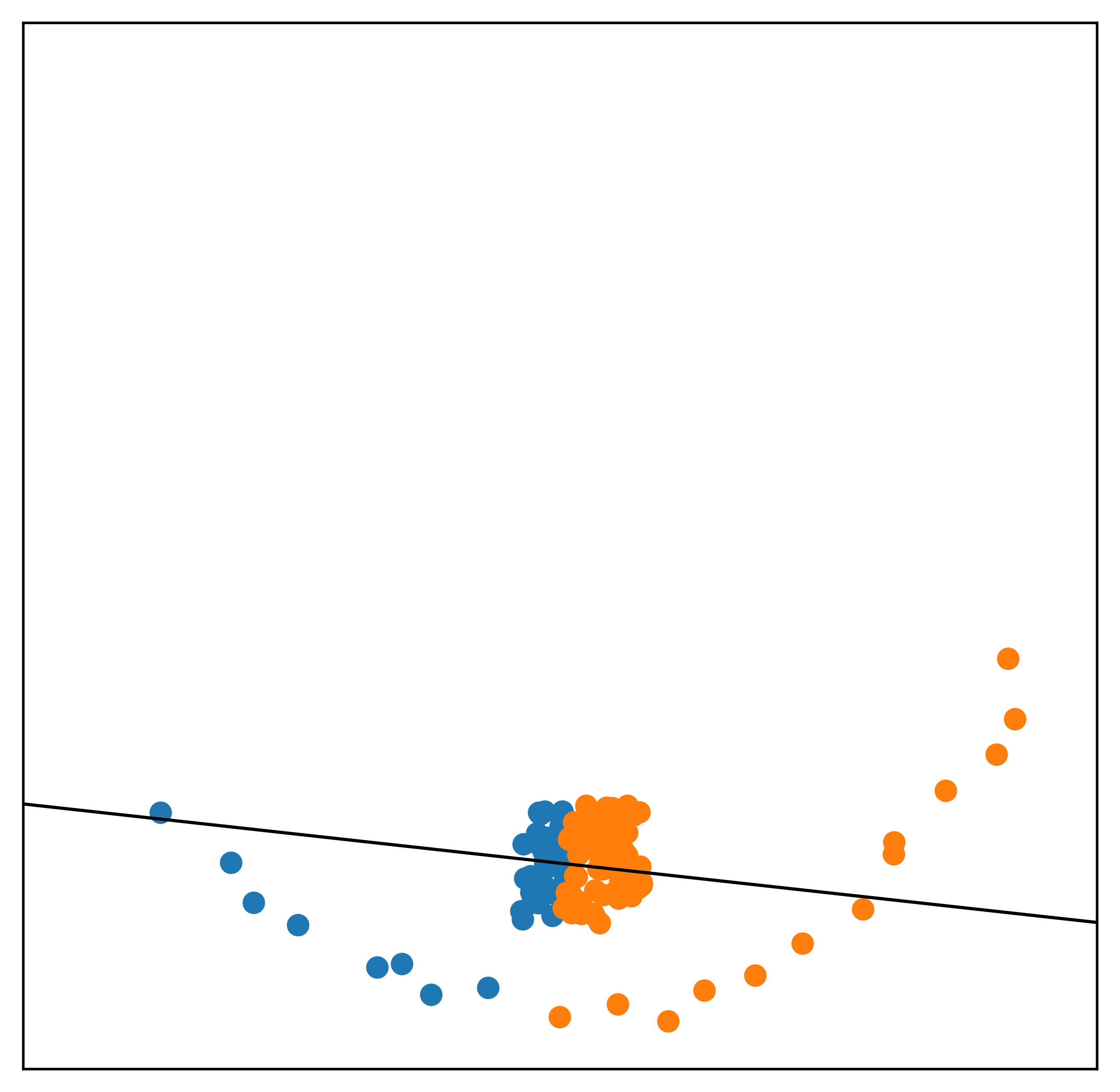}
	\caption{(left) Sampling 100 projection directions with maximum orthogonality between them; (right) splitting points along the direction with maximum dispersion. (Best viewed in color)}
	\label{Fig:sample-directions}
\end{figure}

\begin{figure}[h]
	\centering
	\includegraphics[width=\textwidth]{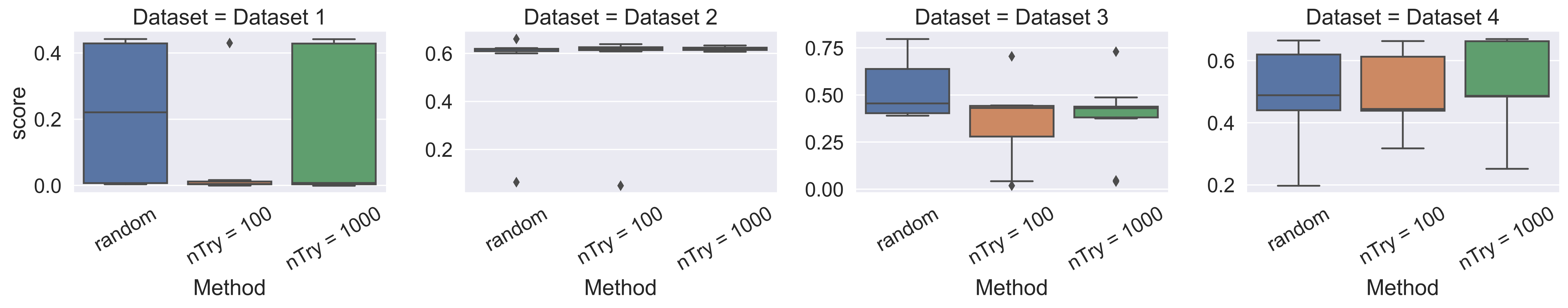}  
	\includegraphics[width=\textwidth]{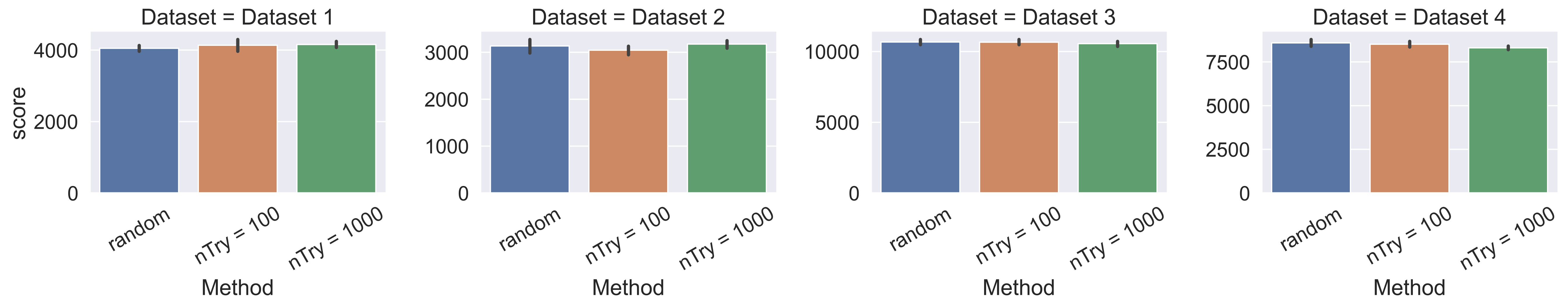}
	\caption{Experiments with synthetic datasets; \texttt{random} represents picking one random projection direction, \texttt{$nTry=100$} and \texttt{$nTry=1000$} is the number of sampled directions; (top) ARI scores for 10 runs, (bottom) number of total pairs. (Best viewed in color)}
	\label{Fig:Results-synthetic-04}
\end{figure}

\begin{figure}[h]
	\centering
	\includegraphics[width=\textwidth]{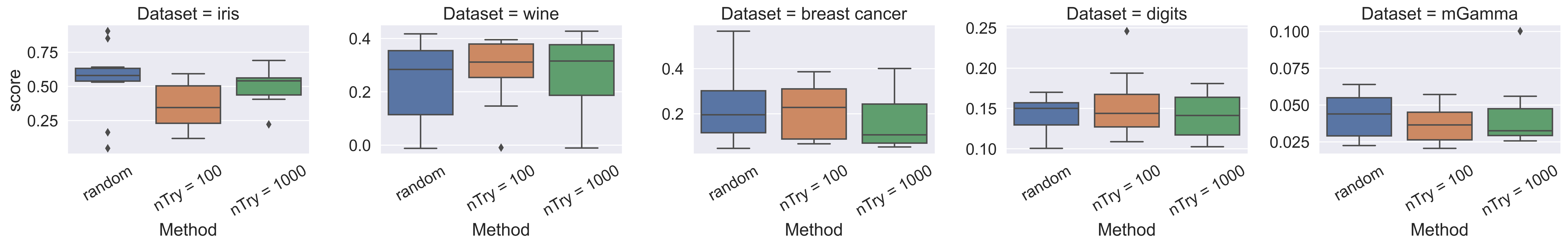}
	\includegraphics[width=\textwidth]{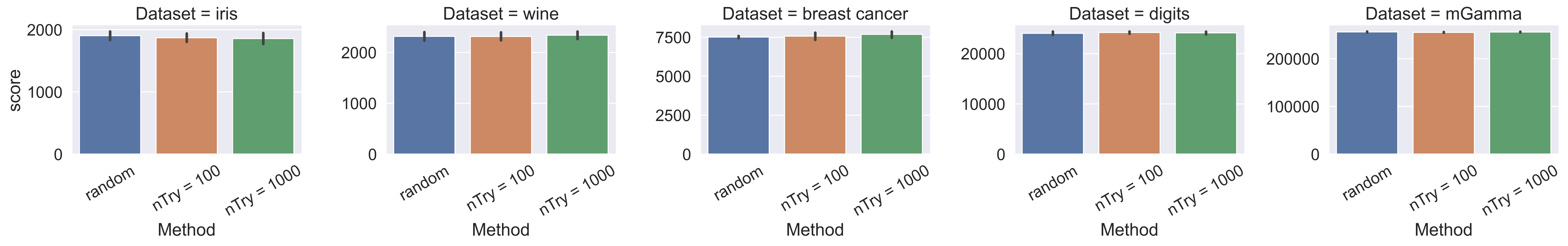}
	\caption{Experiments with real datasets; \texttt{random} represents picking one random projection direction, \texttt{$nTry=100$} and \texttt{$nTry=1000$} is the number of sampled directions; (top) ARI scores for 10 runs, (bottom) number of total pairs. (Best viewed in color)}
	\label{Fig:Results-real-04}
\end{figure}

In a final experiment, we measured the accuracy differences between a choosing random projection direction against an ideal projection direction. As there are infinite number of projection directions, we instead sample up to 1000 different directions uniformly in the unit sphere, and then pick the best performing among those (see Fig.\ \ref{Fig:sample-directions}). For the tested datasets, we compared the best performing direction against the random direction. We found no significant difference among mean of those 100 or 1000 samples with the random vector as shown in Fig.\ \ref{Fig:Results-synthetic-04} and Fig.\ \ref{Fig:Results-real-04}. Our finding is supported by the recent efforts in the literature \cite{Keivani2021Random} to limit the number of random directions to make rpTrees more storage efficient.

\section{Conclusion}
\label{Conclusion}
The conventional way for graph clustering involves the expensive step of eigen decomposition. SpectralNet presents a suitable alternative that does not use eigen decomposition. Instead the embedding is achieved by neural networks.

The similarity metric that was used in SpectralNet was a distance metric for $k$-nearest neighbor graph. This approach restricts points from being paired with further neighbors because $k$ is fixed. A similarity metric based on random projection trees (rpTrees) eases this restriction and allows points to pair with all points falling in the same leaf node. The proposed similarity metric improved the clustering performance on the tested datasets.

There are number of parameters associated with rpTree similarity metric. Parameters like the minimum number of points in a leaf node $n_0$, and how to select the projection direction to split the points. After running experiments while varying these parameters, we found that rpTrees parameters have a limited effect on the clustering performance. So we recommend keeping the number of points in a leaf node $n_0$ in order of $log(n)$. Also, it is more efficient to project the points onto a random direction, instead of trying to find the direction with the maximum dispersion. We conclude that random projection trees (rpTrees) can be used as a similarity metric, where they are applied efficiently as described in this paper.

This work can be extended by changing how the pairwise similarity is computed inside the Siamese net. Currently it is done via a heat kernel. Also, one could use other random projection methods such as random projection forests (rpForest) or rpTrees with reduced space complexity. It would be beneficial for the field to see how these space-partitioning trees perform with clustering in deep networks.





\begin{singlespace}
\bibliographystyle{elsarticle-num-names}
\bibliography{mybibfile}

\begin{thebibliography}{33}
\expandafter\ifx\csname natexlab\endcsname\relax\def\natexlab#1{#1}\fi
\providecommand{\url}[1]{\texttt{#1}}
\providecommand{\href}[2]{#2}
\providecommand{\path}[1]{#1}
\providecommand{\DOIprefix}{doi:}
\providecommand{\ArXivprefix}{arXiv:}
\providecommand{\URLprefix}{URL: }
\providecommand{\Pubmedprefix}{pmid:}
\providecommand{\doi}[1]{\href{http://dx.doi.org/#1}{\path{#1}}}
\providecommand{\Pubmed}[1]{\href{pmid:#1}{\path{#1}}}
\providecommand{\bibinfo}[2]{#2}
\ifx\xfnm\relax \def\xfnm[#1]{\unskip,\space#1}\fi
\bibitem[{Shaham et~al.(2018)Shaham, Stanton, Li, Nadler, Basri, and
  Kluger}]{shaham2018spectralnet}
\bibinfo{author}{U.~Shaham}, \bibinfo{author}{K.~Stanton},
  \bibinfo{author}{H.~Li}, \bibinfo{author}{B.~Nadler},
  \bibinfo{author}{R.~Basri}, \bibinfo{author}{Y.~Kluger},
\newblock \bibinfo{title}{Spectralnet: Spectral clustering using deep neural
  networks},
\newblock in: \bibinfo{booktitle}{6th International Conference on Learning
  Representations, ICLR 2018 - Conference Track Proceedings},
  \bibinfo{year}{2018}. \URLprefix
  \url{https://www.scopus.com/inward/record.uri?eid=2-s2.0-85083950872&partnerID=40&md5=a1d859bb0bc2080faa2ee428a30201b1}.
\bibitem[{Bromley et~al.(1993)Bromley, Guyon, LeCun, S\"{a}ckinger, and
  Shah}]{Bromley1993Signature}
\bibinfo{author}{J.~Bromley}, \bibinfo{author}{I.~Guyon},
  \bibinfo{author}{Y.~LeCun}, \bibinfo{author}{E.~S\"{a}ckinger},
  \bibinfo{author}{R.~Shah},
\newblock \bibinfo{title}{Signature verification using a “siamese” time
  delay neural network},
\newblock in: \bibinfo{booktitle}{Proceedings of the 6th International
  Conference on Neural Information Processing Systems}, NIPS’93,
  \bibinfo{publisher}{Morgan Kaufmann Publishers Inc.}, \bibinfo{address}{San
  Francisco, CA, USA}, \bibinfo{year}{1993}, p. \bibinfo{pages}{737–744}.
\bibitem[{Dasgupta and Freund(2008)}]{Dasgupta2008Random}
\bibinfo{author}{S.~Dasgupta}, \bibinfo{author}{Y.~Freund},
\newblock \bibinfo{title}{Random projection trees and low dimensional
  manifolds},
\newblock in: \bibinfo{booktitle}{Proceedings of the Fortieth Annual ACM
  Symposium on Theory of Computing}, STOC '08, \bibinfo{publisher}{Association
  for Computing Machinery}, \bibinfo{address}{New York, NY, USA},
  \bibinfo{year}{2008}, p. \bibinfo{pages}{537–546}. \URLprefix
  \url{https://doi.org/10.1145/1374376.1374452}.
  \DOIprefix\doi{10.1145/1374376.1374452}.
\bibitem[{Freund et~al.(2008)Freund, Dasgupta, Kabra, and
  Verma}]{Freund2008Learning}
\bibinfo{author}{Y.~Freund}, \bibinfo{author}{S.~Dasgupta},
  \bibinfo{author}{M.~Kabra}, \bibinfo{author}{N.~Verma},
\newblock \bibinfo{title}{Learning the structure of manifolds using random
  projections},
\newblock in: \bibinfo{editor}{J.~Platt}, \bibinfo{editor}{D.~Koller},
  \bibinfo{editor}{Y.~Singer}, \bibinfo{editor}{S.~Roweis} (Eds.),
  \bibinfo{booktitle}{Advances in Neural Information Processing Systems},
  volume~\bibinfo{volume}{20}, \bibinfo{publisher}{Curran Associates, Inc.},
  \bibinfo{year}{2008}. \URLprefix
  \url{https://proceedings.neurips.cc/paper/2007/file/9fc3d7152ba9336a670e36d0ed79bc43-Paper.pdf}.
\bibitem[{Song et~al.(2022)Song, Li, Cai, Yang, Yang, and Liu}]{Song2022Survey}
\bibinfo{author}{X.~Song}, \bibinfo{author}{J.~Li}, \bibinfo{author}{T.~Cai},
  \bibinfo{author}{S.~Yang}, \bibinfo{author}{T.~Yang},
  \bibinfo{author}{C.~Liu},
\newblock \bibinfo{title}{A survey on deep learning based knowledge tracing},
\newblock \bibinfo{journal}{Knowledge-Based Systems} \bibinfo{volume}{258}
  (\bibinfo{year}{2022}) \bibinfo{pages}{110036}. \URLprefix
  \url{https://www.sciencedirect.com/science/article/pii/S0950705122011297}.
  \DOIprefix\doi{https://doi.org/10.1016/j.knosys.2022.110036}.
\bibitem[{Zhou et~al.(2020)Zhou, Huang, Hu, and He}]{ZHOU2020Modeling}
\bibinfo{author}{J.~Zhou}, \bibinfo{author}{J.~X. Huang},
  \bibinfo{author}{Q.~V. Hu}, \bibinfo{author}{L.~He},
\newblock \bibinfo{title}{{SK-GCN}: Modeling syntax and knowledge via graph
  convolutional network for aspect-level sentiment classification},
\newblock \bibinfo{journal}{Knowledge-Based Systems} \bibinfo{volume}{205}
  (\bibinfo{year}{2020}) \bibinfo{pages}{106292}.
  \DOIprefix\doi{https://doi.org/10.1016/j.knosys.2020.106292}.
\bibitem[{Kipf and Welling(2017)}]{kipf2017semi}
\bibinfo{author}{T.~N. Kipf}, \bibinfo{author}{M.~Welling},
\newblock \bibinfo{title}{Semi-supervised classification with graph
  convolutional networks},
\newblock in: \bibinfo{booktitle}{International Conference on Learning
  Representations (ICLR)}, \bibinfo{year}{2017}.
\bibitem[{Franceschi et~al.(2019)Franceschi, Niepert, Pontil, and
  He}]{franceschi2019learning}
\bibinfo{author}{L.~Franceschi}, \bibinfo{author}{M.~Niepert},
  \bibinfo{author}{M.~Pontil}, \bibinfo{author}{X.~He},
\newblock \bibinfo{title}{Learning discrete structures for graph neural
  networks},
\newblock in: \bibinfo{booktitle}{Proceedings of the 36th International
  Conference on Machine Learning}, \bibinfo{year}{2019}.
\bibitem[{Yang et~al.(2022)Yang, Cai, Cai, Song, Jiang, Li, and
  Li}]{Yang2022Robust}
\bibinfo{author}{S.~Yang}, \bibinfo{author}{B.~Cai}, \bibinfo{author}{T.~Cai},
  \bibinfo{author}{X.~Song}, \bibinfo{author}{J.~Jiang},
  \bibinfo{author}{B.~Li}, \bibinfo{author}{J.~Li},
\newblock \bibinfo{title}{Robust cross-network node classification via
  constrained graph mutual information},
\newblock \bibinfo{journal}{Knowledge-Based Systems} \bibinfo{volume}{257}
  (\bibinfo{year}{2022}) \bibinfo{pages}{109852}. \URLprefix
  \url{https://www.sciencedirect.com/science/article/pii/S0950705122009455}.
  \DOIprefix\doi{https://doi.org/10.1016/j.knosys.2022.109852}.
\bibitem[{Yang et~al.(2021)Yang, Verma, Cai, Jiang, Yu, Chen, and
  Yu}]{Yang2021Variational}
\bibinfo{author}{S.~Yang}, \bibinfo{author}{S.~Verma},
  \bibinfo{author}{B.~Cai}, \bibinfo{author}{J.~Jiang},
  \bibinfo{author}{K.~Yu}, \bibinfo{author}{F.~Chen}, \bibinfo{author}{S.~Yu},
  \bibinfo{title}{Variational co-embedding learning for attributed network
  clustering}, \bibinfo{year}{2021}.
  \DOIprefix\doi{https://doi.org/10.48550/ARXIV.2104.07295}.
\bibitem[{Wang et~al.(2022)Wang, Chang, Fu, and Zhao}]{WANG2022Learning}
\bibinfo{author}{Y.~Wang}, \bibinfo{author}{D.~Chang}, \bibinfo{author}{Z.~Fu},
  \bibinfo{author}{Y.~Zhao},
\newblock \bibinfo{title}{Learning a bi-directional discriminative
  representation for deep clustering},
\newblock \bibinfo{journal}{Pattern Recognition}  (\bibinfo{year}{2022})
  \bibinfo{pages}{109237}.
  \DOIprefix\doi{https://doi.org/10.1016/j.patcog.2022.109237}.
\bibitem[{McInnes et~al.(2018)McInnes, Healy, and Melville}]{McInnes2018UMAP}
\bibinfo{author}{L.~McInnes}, \bibinfo{author}{J.~Healy},
  \bibinfo{author}{J.~Melville}, \bibinfo{title}{Umap: Uniform manifold
  approximation and projection for dimension reduction}, \bibinfo{year}{2018}.
  \DOIprefix\doi{https://doi.org/10.48550/ARXIV.1802.03426}.
\bibitem[{Affeldt et~al.(2020)Affeldt, Labiod, and Nadif}]{Affeldt2020Spectral}
\bibinfo{author}{S.~Affeldt}, \bibinfo{author}{L.~Labiod},
  \bibinfo{author}{M.~Nadif},
\newblock \bibinfo{title}{Spectral clustering via ensemble deep autoencoder
  learning (sc-edae)},
\newblock \bibinfo{journal}{Pattern Recognition} \bibinfo{volume}{108}
  (\bibinfo{year}{2020}) \bibinfo{pages}{107522}. \URLprefix
  \url{https://www.sciencedirect.com/science/article/pii/S0031320320303253}.
  \DOIprefix\doi{https://doi.org/10.1016/j.patcog.2020.107522}.
\bibitem[{Huang et~al.(2019)Huang, Ota, Dong, and
  Li}]{Huang2019MultiSpectralNet}
\bibinfo{author}{S.~Huang}, \bibinfo{author}{K.~Ota},
  \bibinfo{author}{M.~Dong}, \bibinfo{author}{F.~Li},
\newblock \bibinfo{title}{Multispectralnet: Spectral clustering using deep
  neural network for multi-view data},
\newblock \bibinfo{journal}{IEEE Transactions on Computational Social Systems}
  \bibinfo{volume}{6} (\bibinfo{year}{2019}) \bibinfo{pages}{749--760}.
  \DOIprefix\doi{10.1109/TCSS.2019.2926450}.
\bibitem[{Wada et~al.(2019)Wada, Miyamoto, Nakagama, Andéol, Kumagai, and
  Kanamori}]{Wada2019Spectral}
\bibinfo{author}{Y.~Wada}, \bibinfo{author}{S.~Miyamoto},
  \bibinfo{author}{T.~Nakagama}, \bibinfo{author}{L.~Andéol},
  \bibinfo{author}{W.~Kumagai}, \bibinfo{author}{T.~Kanamori},
\newblock \bibinfo{title}{Spectral embedded deep clustering},
\newblock \bibinfo{journal}{Entropy} \bibinfo{volume}{21}
  (\bibinfo{year}{2019}). \URLprefix
  \url{https://www.mdpi.com/1099-4300/21/8/795}.
  \DOIprefix\doi{10.3390/e21080795}.
\bibitem[{Zhang et~al.(2021)Zhang, Liu, Wu, Zhang, and Liu}]{ZHANG2021Spectral}
\bibinfo{author}{X.~Zhang}, \bibinfo{author}{H.~Liu}, \bibinfo{author}{X.-M.
  Wu}, \bibinfo{author}{X.~Zhang}, \bibinfo{author}{X.~Liu},
\newblock \bibinfo{title}{Spectral embedding network for attributed graph
  clustering},
\newblock \bibinfo{journal}{Neural Networks} \bibinfo{volume}{142}
  (\bibinfo{year}{2021}) \bibinfo{pages}{388--396}. \URLprefix
  \url{https://www.sciencedirect.com/science/article/pii/S0893608021002227}.
  \DOIprefix\doi{https://doi.org/10.1016/j.neunet.2021.05.026}.
\bibitem[{Jarvis and Patrick(1973)}]{Jarvis1973Clustering}
\bibinfo{author}{R.~Jarvis}, \bibinfo{author}{E.~Patrick},
\newblock \bibinfo{title}{Clustering using a similarity measure based on shared
  near neighbors},
\newblock \bibinfo{journal}{IEEE Transactions on Computers}
  \bibinfo{volume}{C-22} (\bibinfo{year}{1973}) \bibinfo{pages}{1025--1034}.
  \DOIprefix\doi{10.1109/T-C.1973.223640}.
\bibitem[{Wen et~al.(2020)Wen, Zhu, and Zheng}]{Wen2020Spectral}
\bibinfo{author}{G.~Wen}, \bibinfo{author}{Y.~Zhu}, \bibinfo{author}{W.~Zheng},
\newblock \bibinfo{title}{Spectral representation learning for one-step
  spectral rotation clustering},
\newblock \bibinfo{journal}{Neurocomputing} \bibinfo{volume}{406}
  (\bibinfo{year}{2020}) \bibinfo{pages}{361--370}. \URLprefix
  \url{https://www.sciencedirect.com/science/article/pii/S0925231220303477}.
  \DOIprefix\doi{https://doi.org/10.1016/j.neucom.2019.09.108}.
\bibitem[{Zelnik-Manor and Perona(2005)}]{Zelnik2005Self}
\bibinfo{author}{L.~Zelnik-Manor}, \bibinfo{author}{P.~Perona},
\newblock \bibinfo{title}{Self-tuning spectral clustering},
\newblock \bibinfo{journal}{Advances in Neural Information Processing Systems}
  (\bibinfo{year}{2005}) \bibinfo{pages}{1601--1608}.
\bibitem[{Kim et~al.(2021)Kim, Choi, Park, Leung, and Nasridinov}]{Kim2021KNN}
\bibinfo{author}{J.-H. Kim}, \bibinfo{author}{J.-H. Choi},
  \bibinfo{author}{Y.-H. Park}, \bibinfo{author}{C.~K.-S. Leung},
  \bibinfo{author}{A.~Nasridinov},
\newblock \bibinfo{title}{Knn-sc: Novel spectral clustering algorithm using
  k-nearest neighbors},
\newblock \bibinfo{journal}{IEEE Access} \bibinfo{volume}{9}
  (\bibinfo{year}{2021}) \bibinfo{pages}{152616--152627}.
  \DOIprefix\doi{10.1109/ACCESS.2021.3126854}.
\bibitem[{Ester et~al.(1996)Ester, Kriegel, Sander, and Xu}]{Ester1996Density}
\bibinfo{author}{M.~Ester}, \bibinfo{author}{H.-P. Kriegel},
  \bibinfo{author}{J.~Sander}, \bibinfo{author}{X.~Xu},
\newblock \bibinfo{title}{A density-based algorithm for discovering clusters in
  large spatial databases with noise},
\newblock in: \bibinfo{booktitle}{Proceedings of the Second International
  Conference on Knowledge Discovery and Data Mining}, KDD'96,
  \bibinfo{publisher}{AAAI Press}, \bibinfo{year}{1996}, p.
  \bibinfo{pages}{226–231}.
\bibitem[{Yan et~al.(2009)Yan, Huang, and Jordan}]{Yan2009Fast}
\bibinfo{author}{D.~Yan}, \bibinfo{author}{L.~Huang}, \bibinfo{author}{M.~I.
  Jordan},
\newblock \bibinfo{title}{Fast approximate spectral clustering},
\newblock in: \bibinfo{booktitle}{Proceedings of the 15th ACM SIGKDD
  International Conference on Knowledge Discovery and Data Mining}, KDD '09,
  \bibinfo{publisher}{Association for Computing Machinery},
  \bibinfo{address}{New York, NY, USA}, \bibinfo{year}{2009}, p.
  \bibinfo{pages}{907–916}. \URLprefix
  \url{https://doi.org/10.1145/1557019.1557118}.
  \DOIprefix\doi{10.1145/1557019.1557118}.
\bibitem[{Yan et~al.(2019)Yan, Gu, Xu, and Qin}]{Yan2019Similarity}
\bibinfo{author}{D.~Yan}, \bibinfo{author}{S.~Gu}, \bibinfo{author}{Y.~Xu},
  \bibinfo{author}{Z.~Qin},
\newblock \bibinfo{title}{Similarity kernel and clustering via random
  projection forests},
\newblock \bibinfo{journal}{CoRR} \bibinfo{volume}{abs/1908.10506}
  (\bibinfo{year}{2019}). \URLprefix \url{http://arxiv.org/abs/1908.10506}.
  \href{http://arxiv.org/abs/1908.10506}{{\tt arXiv:1908.10506}}.
\bibitem[{Gilbert(1961)}]{Gilbert1961Random}
\bibinfo{author}{E.~N. Gilbert},
\newblock \bibinfo{title}{Random plane networks},
\newblock \bibinfo{journal}{Journal of the Society for Industrial and Applied
  Mathematics} \bibinfo{volume}{9} (\bibinfo{year}{1961})
  \bibinfo{pages}{533--543}. \DOIprefix\doi{10.1137/0109045}.
\bibitem[{Barthelemy(2017)}]{barthelemy2017morphogenesis}
\bibinfo{author}{M.~Barthelemy}, \bibinfo{title}{Morphogenesis of Spatial
  Networks}, Lecture Notes in Morphogenesis, \bibinfo{publisher}{Springer
  International Publishing}, \bibinfo{year}{2017}.
\bibitem[{Hubert and Arabie(1985)}]{Hubert1985Comparing}
\bibinfo{author}{L.~Hubert}, \bibinfo{author}{P.~Arabie},
\newblock \bibinfo{title}{Comparing partitions},
\newblock \bibinfo{journal}{Journal of Classification} \bibinfo{volume}{2}
  (\bibinfo{year}{1985}) \bibinfo{pages}{193--218}. \URLprefix
  \url{https://www.scopus.com/inward/record.uri?eid=2-s2.0-0000008146&doi=10.1007%2fBF01908075&partnerID=40&md5=bd03cf70caee7de0ccf3c0dd431b97ca}.
  \DOIprefix\doi{10.1007/BF01908075}.
\bibitem[{Pedregosa et~al.(2011)Pedregosa, Varoquaux, Gramfort, Michel,
  Thirion, Grisel, Blondel, Prettenhofer, Weiss, Dubourg, Vanderplas, Passos,
  Cournapeau, Brucher, Perrot, and Duchesnay}]{scikit-learn}
\bibinfo{author}{F.~Pedregosa}, \bibinfo{author}{G.~Varoquaux},
  \bibinfo{author}{A.~Gramfort}, \bibinfo{author}{V.~Michel},
  \bibinfo{author}{B.~Thirion}, \bibinfo{author}{O.~Grisel},
  \bibinfo{author}{M.~Blondel}, \bibinfo{author}{P.~Prettenhofer},
  \bibinfo{author}{R.~Weiss}, \bibinfo{author}{V.~Dubourg},
  \bibinfo{author}{J.~Vanderplas}, \bibinfo{author}{A.~Passos},
  \bibinfo{author}{D.~Cournapeau}, \bibinfo{author}{M.~Brucher},
  \bibinfo{author}{M.~Perrot}, \bibinfo{author}{E.~Duchesnay},
\newblock \bibinfo{title}{Scikit-learn: Machine learning in {P}ython},
\newblock \bibinfo{journal}{Journal of Machine Learning Research}
  \bibinfo{volume}{12} (\bibinfo{year}{2011}) \bibinfo{pages}{2825--2830}.
\bibitem[{Buitinck et~al.(2013)Buitinck, Louppe, Blondel, Pedregosa, Mueller,
  Grisel, Niculae, Prettenhofer, Gramfort, Grobler, Layton, VanderPlas, Joly,
  Holt, and Varoquaux}]{sklearn_api}
\bibinfo{author}{L.~Buitinck}, \bibinfo{author}{G.~Louppe},
  \bibinfo{author}{M.~Blondel}, \bibinfo{author}{F.~Pedregosa},
  \bibinfo{author}{A.~Mueller}, \bibinfo{author}{O.~Grisel},
  \bibinfo{author}{V.~Niculae}, \bibinfo{author}{P.~Prettenhofer},
  \bibinfo{author}{A.~Gramfort}, \bibinfo{author}{J.~Grobler},
  \bibinfo{author}{R.~Layton}, \bibinfo{author}{J.~VanderPlas},
  \bibinfo{author}{A.~Joly}, \bibinfo{author}{B.~Holt},
  \bibinfo{author}{G.~Varoquaux},
\newblock \bibinfo{title}{{API} design for machine learning software:
  experiences from the scikit-learn project},
\newblock in: \bibinfo{booktitle}{ECML PKDD Workshop: Languages for Data Mining
  and Machine Learning}, \bibinfo{year}{2013}, pp. \bibinfo{pages}{108--122}.
\bibitem[{Dua and Graff(2017)}]{Dua2019UCI}
\bibinfo{author}{D.~Dua}, \bibinfo{author}{C.~Graff}, \bibinfo{title}{{UCI}
  machine learning repository}, \bibinfo{year}{2017}. \URLprefix
  \url{http://archive.ics.uci.edu/ml}.
\bibitem[{Alshammari et~al.(2021)Alshammari, Stavrakakis, and
  Takatsuka}]{Alshammari2021Refining}
\bibinfo{author}{M.~Alshammari}, \bibinfo{author}{J.~Stavrakakis},
  \bibinfo{author}{M.~Takatsuka},
\newblock \bibinfo{title}{Refining a k-nearest neighbor graph for a
  computationally efficient spectral clustering},
\newblock \bibinfo{journal}{Pattern Recognition} \bibinfo{volume}{114}
  (\bibinfo{year}{2021}) \bibinfo{pages}{107869}. \URLprefix
  \url{https://www.sciencedirect.com/science/article/pii/S003132032100056X}.
  \DOIprefix\doi{https://doi.org/10.1016/j.patcog.2021.107869}.
\bibitem[{Ram and Sinha(2019)}]{Ram2019Revisiting}
\bibinfo{author}{P.~Ram}, \bibinfo{author}{K.~Sinha},
\newblock \bibinfo{title}{Revisiting kd-tree for nearest neighbor search},
\newblock \bibinfo{publisher}{Association for Computing Machinery},
  \bibinfo{address}{New York, NY, USA}, \bibinfo{year}{2019}, p.
  \bibinfo{pages}{1378–1388}.
  \DOIprefix\doi{https://doi.org/10.1145/3292500.3330875}.
\bibitem[{Yan et~al.(2021)Yan, Wang, Wang, Wang, and Li}]{Yan2021Nearest}
\bibinfo{author}{D.~Yan}, \bibinfo{author}{Y.~Wang}, \bibinfo{author}{J.~Wang},
  \bibinfo{author}{H.~Wang}, \bibinfo{author}{Z.~Li},
\newblock \bibinfo{title}{K-nearest neighbor search by random projection
  forests},
\newblock \bibinfo{journal}{IEEE Transactions on Big Data} \bibinfo{volume}{7}
  (\bibinfo{year}{2021}) \bibinfo{pages}{147--157}.
  \DOIprefix\doi{10.1109/TBDATA.2019.2908178}.
\bibitem[{Keivani and Sinha(2021)}]{Keivani2021Random}
\bibinfo{author}{O.~Keivani}, \bibinfo{author}{K.~Sinha},
\newblock \bibinfo{title}{Random projection-based auxiliary information can
  improve tree-based nearest neighbor search},
\newblock \bibinfo{journal}{Information Sciences} \bibinfo{volume}{546}
  (\bibinfo{year}{2021}) \bibinfo{pages}{526--542}.
  \DOIprefix\doi{https://doi.org/10.1016/j.ins.2020.08.054}.

\end{thebibliography}
\end{singlespace}

\end{document}